%% file: acl_latex.tex
\pdfoutput=1

\documentclass[11pt]{article}

\usepackage[dvipsnames]{xcolor}

\usepackage[]{acl}

\usepackage{times}
\usepackage{latexsym}

\usepackage[T1]{fontenc}

\usepackage[utf8]{inputenc}

\usepackage{microtype}

\usepackage{inconsolata}

\usepackage{times}
\usepackage{graphicx}
\usepackage{multirow}
\usepackage{latexsym}
\usepackage[normalem]{ulem}

\usepackage{paralist} 
\usepackage{enumitem}
\usepackage{array}
\usepackage{booktabs}
\usepackage{pifont}
\usepackage{amssymb}
\usepackage{MnSymbol}
\newlist{todolist}{itemize}{2}
\setlist[todolist]{label=$\square$}

\usepackage{csquotes}
\usepackage{soul, color}

\newcommand{\para}[1]{\vspace{4pt}\noindent\textbf{#1}~}
\definecolor{vanilla}{rgb}{0.95, 0.9, 0.67}
\DeclareRobustCommand{\hlcolor}[1]{{\sethlcolor{vanilla}\hl{#1}}}

\title{``\textit{One-Size-Fits-All}''? Examining Expectations around What Constitute ``Fair'' or ``Good'' NLG System Behaviors} 
\author{Li Lucy\textsuperscript{2} \, Su Lin Blodgett\textsuperscript{1} \, Milad Shokouhi\textsuperscript{1} \, Hanna Wallach\textsuperscript{1}\, Alexandra Olteanu\textsuperscript{1}\\ 
\textsuperscript{1}Microsoft Research \\
\textsuperscript{2}University of California, Berkeley \\ 
\texttt{lucy3\_li@berkeley.edu} \\
\texttt{\{sulin.blodgett,milads,wallach,alexandra.olteanu\}@microsoft.com}
}

\begin{document}
\maketitle

\begin{abstract}

\input{paper-files/sections/0_abstract}
\end{abstract}

\input{paper-files/sections/1_introduction}

\input{paper-files/sections/2_fair_lang_id}

\section{Case Studies}

\input{paper-files/sections/3_case_studies}

\section{Categories of System Behaviors}\label{observed_section}

\input{paper-files/sections/4_observed_behaviors}

\section{Expectations of System Behaviors}\label{expected}

\input{paper-files/sections/5_task_design}

\input{paper-files/sections/5.1_expected_behaviors}

\section{Conclusion}\label{conclusions}

\input{paper-files/sections/6_conclusion}

\input{paper-files/sections/7-8_limitations_ethics}

\bibliography{custom}

\appendix

\input{paper-files/sections/Appendix_A}

\input{paper-files/sections/Appendix_B}

\input{paper-files/sections/Appendix_C}

\input{paper-files/sections/Appendix_D}

\input{paper-files/sections/Appendix_E}

\end{document}

%% file: paper-files/sections/0_abstract.tex










Fairness-related assumptions about what constitute appropriate NLG system behaviors range from {\em invariance}, where systems are expected to behave identically for social groups, to {\em adaptation}, where behaviors should instead vary across them. To illuminate tensions around invariance and adaptation, we conduct five case studies, in which we perturb different types of identity-related language features (names, roles, locations, dialect, and style) in NLG system inputs. Through these cases studies, we examine people's expectations of system behaviors, and surface potential caveats of these contrasting yet commonly held assumptions. We find that motivations for adaptation include social norms, cultural differences, feature-specific information, and accommodation; in contrast, motivations for invariance include perspectives that favor prescriptivism, view adaptation as unnecessary or too difficult for NLG systems to do appropriately, and are wary of false assumptions. Our findings highlight open challenges around what constitute~``fair'' or ``good'' NLG system behaviors.\looseness=-1

%% file: paper-files/sections/1_introduction.tex
\section{Introduction} 


Natural language generation (NLG) models are used for many downstream applications involving interpersonal communication, such as text completion, ``smart'' reply suggestions, and chatbot assistants \cite{mieczkowski_2021, trajanovski-etal-2021-text, buschek2021impact, liu_ai_email2022}. At the same time, there are  growing concerns that NLG models and the systems that incorporate them may reproduce or exacerbate biases, causing harms that affect subsets of people \cite{robertson2021can, amershi_2019, hancock_aimc_2020, jakesch_2019, sheng-etal-2021-societal}. Addressing these concerns requires us to be able to specify what model or system behaviors are ``fair,'' which may extend beyond behavior patterns within the scope of common or existing definitions of fairness.\looseness=-1

More generally, the task of specifying desirable or ``good'' NLG model or system behaviors---of which specifying ``fair'' behaviors is one example---is non-trivial. A key challenge is that concepts like ``good'' and ``fair'' are {\em essentially contested constructs}~\cite{jacobs2021measurement}---i.e., they have multiple context-specific, and sometimes even conflicting, definitions. 
To illustrate this challenge, we surface tensions between two commonly held fairness-related assumptions: {\em invariance}, where systems are expected to behave identically for social groups, and {\em adaptation}, where instead system behaviors are expected to vary across social groups. 

On the side of {\em invariance}, definitions of fairness assume social groups should be treated the same \cite{benthall_haynes_2019, smith2021hi, elazar-goldberg-2018-adversarial,romanov-etal-2019-whats}. However, approaches that treat social labels as interchangeable may not account for valid differences between groups, mediated by historical, political, and social contexts \cite{hanna_race_2020, mostafazadeh-davani-etal-2021-improving}. Invariance can thus lead to alienation \cite{garg2019counterfactual}, factuality issues \cite{qian-etal-2022-perturbation}, and language homogenization \cite{hancock_aimc_2020, hovy-etal-2020-sound}. On the side of {\em adaptation}, some people favor personalization or customization based on social identity \cite{salewski2023incontext, flek-2020-returning, dudy-etal-2021-refocusing, suriyakumar2022personalization,jin-etal-2022-deep}. However, this can lead to stereotyping, unwanted assumptions, language~appropriation, and offensive responses. \looseness=-1

To initiate a discussion around invariance versus adaptation in the context of NLG models and systems, we use \textit{identity-related language features} to both observe actual NLG system behaviors and examine people's expectations of them. 
We present five case studies that empirically examine system behaviors in the presence of several types of English \textit{language features} that are associated with \textit{social identity}: names, roles, locations, dialect, and style. Focusing on ``smart'' reply suggestions as an illustrative downstream application, these case studies surface potential fairness-related harms, such as quality-of-service and representational harms, arising from various NLG system behaviors \cite{crawford2017trouble,blodgett-etal-2020-language,bird2020fairlearn}. Each case study has two parts: one part in which we use grounded theory methods to categorize observed differences in system behaviors, and another part in which we design crowd experiments to examine people's expectations of system behaviors. We focus on two research questions:\looseness=-1
\begin{compactenum}
    \item[\textbf{RQ1}:] What differences in {\em system behaviors} do we observe when we vary identity-related language features in NLG system inputs? 
    \item[\textbf{RQ2}:] How do {\em people's expectations} of system behaviors vary when we vary identity-related language features in NLG system inputs? 
\end{compactenum}
Our findings surface tensions between whether NLG systems should be \textit{invariant} to identity-related language features or \textit{adapt} based on them, highlighting open challenges around what constitute ``fair'' or ``good'' NLG system behaviors.\looseness=-1

%% file: paper-files/sections/2_fair_lang_id.tex
\section{Fairness, language, \& identity}\label{background}


Evaluating NLG systems is not a straightforward endeavor in practice. Most fairness measurements center around demographic attributes, such as race or gender. However, there are significant legal and practical barriers to acquiring demographic information about users \cite{andrus2021we, holstein2019improving}, and this scarcity of information has led to the use of linguistic proxies, correlates, and markers \cite{tan-etal-2021-reliability, NEURIPS2020_07fc15c9}. Our study similarly adopts this paradigm, operationalizing identity using only language features. Evaluating NLG systems using such features relies on many under-examined assumptions, especially around how NLG systems should respond to them.\looseness=-1

In sociolinguistics, language is a performance of \textit{social identity}, which extends beyond demographic attributes and includes membership in many types of social groups.
We draw on this broad notion of social identity, since it provides a more comprehensive conceptualization of people's relationships with language. 
The use of language features in evaluating NLG systems is limited by the lack of a one-to-one mapping between language and identity (\S\ref{sec:limit}). Concepts such as race and gender are also social constructs that encompass multiple definitions \cite{hanna_race_2020,cao-daume-iii-2020-toward,benthall_haynes_2019,antoniak-mimno-2021-bad}.
Thus, studies that use language features tell us how a system responds to these features, rather than how it may respond to specific social groups.\looseness=-1

The features that we use in our case studies fall under the broad categories of {\em references} and {\em variation}. 
Here, we provide background on these features, including examples of their use in the context of fairness and how they relate to identity.\looseness=-1


\para{References} in text can denote specific individuals or social groups (direct and relative references), or concepts and entities connected to identity (associative references). 
Just as humans are sensitive to social connotations of these references \cite{bjorkman2017singular, nosek2002harvesting, Moss-Racusin16474}, algorithmic systems can reproduce these perceptual patterns. Thus, identity-related references have been used to evaluate models and systems for biases and harms \cite{Caliskan183, smith2021hi, kirk2021true, zhao-etal-2018-gender, sheng-etal-2019-woman, smith-etal-2022-im}.\looseness=-2 

Direct references to individuals include proper names (e.g., \textit{Morgan}, \textit{Priyanka}), sometimes supplemented with titles and pronouns. These references can be used to construct identity \cite{pollitt_trans_2021, cila2019s}, and can be implicitly associated with gender, ethnicity, geography, and age \cite{edwards_2009, blevins2015jane}. 
Other references to people indicate their relative positions in the world or membership in social groups. Examples include occupation (\textit{doctor}), familial role (\textit{son}), geographic origin (\textit{American}), and intersectional identities (\textit{Latina}). \looseness=-1

Associative references are non-person entities linked to social groups via shared cultural and community interests. Examples include locations \cite{zhou-etal-2022-richer}, activities \cite{de2019bias}, and topics \cite{sheng-etal-2021-nice}. Though their associations with social groups vary across contexts and domains 
\cite{bamman_2014, herring_paolillo_2006}, these references can affect model and system behaviors in undesirable ways. \looseness=-1 


\para{Linguistic variation,} or different ways of saying similar things, expresses \textit{social meaning}, or information about a speaker's social identity \cite{nguyen-etal-2021-learning}. \textit{Dialects} can be associated with geographic regions, ethnicities (\textit{ethnolects}), or communities (\textit{sociolects}), with code-switching widening the range of variation. Language varieties can also pertain to specific situations (\textit{registers}), 
and speakers adjust their language \textit{style} based on audience and formality \cite{eckert2017phonetics, bell1984language, pavalanathan2015audience}. Variation occurs at many levels of linguistic analysis, from phonological to lexical, though syntactic variation often raises the most stigma \cite{edwards_2009}. English models perform poorly on minoritized varieties \cite[e.g.,][]{ziems-etal-2022-value}, and some NLP practices, like text normalization, can imply one variety is more valid than others \cite{eisenstein-2013-bad}.


%% file: paper-files/sections/3_case_studies.tex

In this section, we describe the features we use to examine observed (\textbf{RQ1}, \S\ref{observed_section}) and expected (\textbf{RQ2}, \S\ref{expected}) NLG system behaviors across five case studies. The first three case studies vary references to entities: direct (names), relative (parental roles), and associative (countries). The last two examine linguistic variation: dialect and style. For brevity, we reference each case study with ``CS'' and its number.\looseness=-1

In each case study, we craft \textit{message templates} covering a variety of speech acts, for which we then perturb identity-related language features (Table~\ref{tbl:case_studies}). We use these messages as inputs for three different NLG systems to uncover categories of observed system behaviors (\textbf{RQ1}, \S\ref{observed_section}). We then use a subset of the perturbed messages to design \textit{vignettes} consisting of a message and a pair of reply options to surface people's expectations of system behaviors (\textbf{RQ2}, \S\ref{expected}). Details about feature selection and all message templates are in Appendices~\ref{appdx:names}--\ref{appdx:style}.\looseness=-1

\para{CS1: Names.}
To address \textbf{RQ1}, we experiment with over 240 first names from \citet{tzioumis2018demographic} as the sender, recipient, or mentioned third party in five message templates used to study reply suggestions~\cite{robertson2021can}, as some system behaviors, e.g., pronoun assumptions, might only emerge when names appear in particular positions. For \textbf{RQ2}, we use messages containing six names (\textit{Reyna}, \textit{Salim}, \textit{Jackie}, \textit{Annie}, \textit{Kalen}, and \textit{Tony}) reflecting different gender associations
(feminine, masculine, neutral) and levels of familiarity for U.S.-based judges. We experiment with these names in the sender position, except when testing for pronoun assumptions. There, we insert names as a mentioned third party, so pronouns in replies could refer to the name and retain coherence. \looseness=-1

\para{CS2: Parental roles.}
This case study compares names to parental terms, 
to highlight references that differ in how they signal someone's identity relative to others. For parental terms, we use \textit{Mom}, \textit{Mommy}, \textit{Dad}, and \textit{Daddy}, and compare these to \textit{Jennifer} and \textit{Michael}, which are popular, gendered names in the U.S. for people of parental age \cite{ssa_names}. We craft five message templates, similar to those in CS1, but more plausible for communication within families. For example, we revise a message template from CS1 about \textit{scheduling a meeting} into a request to \textit{get together}. We again place references in the closings, greetings, or bodies of messages, which correspond to senders, recipients, or mentioned third parties, for \textbf{RQ1}. For \textbf{RQ2}, we place these references in the sender position in all message--reply vignettes except for those used to test for pronoun assumptions.\looseness=-1


\begin{table}[t] 
\centering
\resizebox{\columnwidth}{!}{
\begin{tabular}{l} 
\toprule
\multicolumn{1}{c}{\textbf{References}} \\
\midrule
\textbf{CS1: Names} \\
It’s been a good week. \hlcolor{Annie} got promoted. \\
\textbf{CS2: Parental Roles} \\
It will be a long day. I'll bring snacks for everyone. Best, \hlcolor{Mom} \\
\textbf{CS3: Countries} \\
Next week, I am traveling home to \hlcolor{Serbia}.\\
 \midrule
\multicolumn{1}{c}{\textbf{Variation}} \\
\midrule
\textbf{CS4: African American English} \\
\textit{multiple negation}\\
\hlcolor{Don't} bring \hlcolor{nothing}. I don't need your help in this kitchen.\\
\textit{habitual be}\\
You should totally come to our party, we \hlcolor{be} having so much fun.\\
\textbf{CS5: Informal web text} \\
\textit{expressive elongation}\\
I \hlcolor{realllly} liked the topic of their presentation.\\
\textit{non-standard capitalization}\\
\hlcolor{y}ou guys sounded like you were partying. \hlcolor{d}id you have fun? \\
\textit{complex punctuation}\\
Have a great holiday. I'm out of here\hlcolor{!!!!!!!!!!}\\
\bottomrule
\end{tabular}
}
\caption{Message examples that contain identity-related language features, highlighted, across CS1--5.} 
\label{tbl:case_studies}
\end{table}

\para{CS3: Countries.}
Here, we perturb country names in three message templates: a meeting request, an open-ended question about planned activities, and a travel announcement. For \textbf{RQ1}, we use 226 country names listed by the U.S. Department of State~\cite{gov_countries}. We place countries in positions that signal the sender, recipient, or mentioned third party as from or traveling to the country. For \textbf{RQ2}, we use six countries from three world regions, in pairs that differ in wealth or GDP:\footnote{The ``region'' a country belongs to can vary depending on the source. We select geographically proximate pairs of countries, and derive region labels from those listed by the United Nations: \url{https://unstats.un.org/unsd/methodology/m49/\#geo-regions}. See Appendix~\ref{appdx:locations} for details.\looseness=-1} 
\textit{Italy} and \textit{Serbia} (Southern Europe), \textit{Egypt} and \textit{Eritrea} (Northeast Africa), and \textit{India} and \textit{Afghanistan} (South Asia).  These countries are then used in vignettes where the person associated with the country is the sender.\looseness=-1

\para{CS4: African American English.}
This case study examines features associated with African American English (AAE), which encompasses several dialects that vary based on formality and geography. We examine the presence and absence of two salient syntactic features in messages: multiple negation and habitual \textit{be}. Both features are also used in other English dialects, and often appropriated by non-AAE speakers. Our input message templates are taken from studies that transcribe language from Black AAE speakers \cite{green2002african,Rickford11817}. 
For \textbf{RQ1}, we test six pairs of AAE and General American English (GAE) messages that perturb multiple negation, and six that perturb habitual \textit{be}. For \textbf{RQ2}, we use a subset of two pairs for each feature. \looseness=-1

\para{CS5: Informal web text.}
Here we focus on several features common in informal web text: expressive word lengthening \cite{
kalman2014letter,brody-diakopoulos-2011-cooooooooooooooollllllllllllll}, complex punctuation \cite{rao2010classifying}, and non-standard capitalization \cite{squires2010enregistering}. We craft messages perturbing these features, based on examples found in the Enron email corpus or discussed in prior work on computer-mediated communication \cite{kalman2014letter,brody-diakopoulos-2011-cooooooooooooooollllllllllllll}. They are thus pairs of more or less casual messages. For \textbf{RQ1}, each feature is perturbed in six message pairs, with an additional message that iteratively perturbs and combines all features. For \textbf{RQ2}, we use two message pairs for each feature, along with an additional message that combines them all.\looseness=-1 

%% file: paper-files/sections/4_observed_behaviors.tex
To observe system behaviors (\textbf{RQ1}), we experiment with three NLG systems that pertain to interpersonal communication: a) chat ``smart'' reply suggestions using Google's ML Kit \cite{kannan2016smart}, b) email reply suggestions \cite{deb-etal-2019-diversifying}, and c) dialogue response generation using DialoGPT \cite{zhang-etal-2020-dialogpt}. 
The first two are actively deployed in messaging applications at the time of our study, and retrieve reply suggestions from a pre-curated response space. The third involves open generation with no guardrails or response curation. Thus, these three systems differ in terms of the types of replies they suggest, helping us observe a wider range of system behaviors. 

To identify patterns in reply suggestions across the case studies, we use grounded theory methods, including open and axial coding
\cite{charmaz2006constructing, Muller2014}.  
Three authors coded all unique replies to each message template, which were accompanied by a sampled subset of illustrative messages. They then met to discuss the replies and iterated together to create a coding scheme for~observed differences in system behaviors.\looseness=-1

\para{Coherence.}
Some reply suggestions are less coherent than others, which can potentially lead to quality-of-service harms. Replies that lack coherence include explicit expressions of confusion (e.g., \textit{I'm not sure what you mean by this}) and text that includes implausible, out-of-context information \cite{shwartz-etal-2020-grounded}. Replies may also parrot parts of the message in illogical ways or repeat phrases unnecessarily \cite{fu2021theoretical}. Some replies are semantically incoherent, contradicting~or misinterpreting message content. \looseness=-1

\smallskip
Even when replies are coherent, they can differ in characteristics such as {\em sentiment and affect}, {\em formality}, and {\em complexity}. We describe reply differences using these broad characteristics, acknowledging that some, such as formality and affect, are interconnected with overlapping boundaries.\looseness=-1 

\para{Sentiment and affect.}
We observe that perturbing features in CS1--5 can result in differences in sentiment. Beyond polarity differences in positive answers (e.g., \textit{Sure}) versus negative ones (e.g., \textit{Nope}), we observe differences in sentiment modulated by the inclusion of intensifiers like \textit{so} (e.g., \textit{I'm so happy for him}) and exclamation points. 
Replies can also differ in their affect, including tone, attitude, and emotion. For example, \textit{So proud of you!} might suggest greater familiarity than \textit{So happy for you!}. Replies to some messages are also warmer and more reassuring (e.g., \textit{I understand}) than replies to others (e.g., \textit{Ok, thanks for letting me know}).

\para{Formality.} Replies can also differ in their formality (CS1--5) as indicated by, e.g., emoji use or colloquial wording. Examples include the more informal \textit{Yup} instead of \textit{Yes}, or \textit{I know that feel} instead of \textit{I know, I'm so sorry}. In practice, language can express formality differences in myriad ways, though we did not observe replies that include the informal-web-text features that we perturb in CS5. 


\para{Textual complexity.}
Replies can also differ in their textual complexity, where replies to the same message template can be brief or appended with extra information. Examples of additions include emotive expressions, comments, questions, or actions (e.g., \textit{I did!} versus \textit{I did! Thanks for the followup.}).  
We hypothesize that textual complexity, along with other characteristics such as sentiment, affect, and formality, may impact replies' usabilities for members of different social groups. We discuss possible implications of these textual differences relating to quality-of-service harms in \S\ref{judge_expect}.

\para{Identity-related assumptions.}
In all five case studies, some replies appear to infer characteristics of the sender, recipient, or mentioned third party. Assumptions around gender, age, and relationships are most noticeable in CS1--3, e.g., \textit{I'll ask my wife}. One system generates replies containing gendered pronouns or markers (e.g., \textit{Congrats man!}), while the others avoid this behavior \cite{robertson2021can,vincent2018}. 
Other assumptions relate to interests or behaviors, such as replies that mention alcohol or a specific travel destination (CS2--5). These assumptions vary in their specificity, such as \textit{doing a lot of things} versus \textit{going to the beach}. 
Identity-related assumptions can lead to representational harms, reducing people's agency to define themselves and perpetuating harmful stereotypes. 

\para{Availability of service.}
Deployed systems often implement guardrails, e.g., {\em blocklists}, to prevent undesirable system behaviors~\cite{schlesinger2018let,raffel2020exploring,dodge-etal-2021-documenting,zhou-etal-2022-deconstructing}. 
Indeed, in CS1--3 and CS5, no replies are suggested for some messages. We observe blocking behavior in response to messages that contain the name \textit{Adolph}, more casual language (e.g., \textit{freeeezing} instead of \textit{freezing}), and the vast majority of country names. A lack of replies for messages that contain some identity-related language features can unfairly imply that some social groups have a lesser need for service than others.
%
%

%% file: paper-files/sections/5_task_design.tex
\subsection{Task Design}

\begin{table}[t] 
\centering
\setlength{\tabcolsep}{0.2em}
\tiny
\begin{tabular}{@{}p{0.9cm}p{2.1cm}>{\setlength{\baselineskip}{0.4\baselineskip}}p{3.7cm}c@{}} 
\textbf{Category} & \textbf{Subcategory} & \textbf{Example \textcolor{gray}{baseline reply} / second reply} & \textbf{CS} \\
\midrule
\multirow{ 3}{*}{\bf Coherence}  & expression of confusion & \textcolor{gray}{Yes, all good.} / I'm not sure what you mean by this. & 1--5 \\
& repetition \& parroting & \textcolor{gray}{Sure, I'll come!} / Having so much fun. Having so much fun.  & 4--5 \\
& irrelevant information & \textcolor{gray}{Yes, all good.} / Yes, you left the football game.* & 1--3 \\
& semantically incoherent & \textcolor{gray}{Yes, all good.} / Yes, will do.* & 1--3 \\
\midrule
\multirow{ 4}{*}{\bf Sentiment}  & intensity (increase) & \textcolor{gray}{Yes, all good.} / Yes, all good! & 1--5 \\
& intensity (decrease) & \textcolor{gray}{Yes, all good.} / Yes, okay. & 1--5 \\
& direction (pos $\rightarrow$ neg) & \textcolor{gray}{Yes, all good.} / No, it's not. & 1--5 \\
& more warm affect & \textcolor{gray}{Yes, all good.} / Yes, grateful for your help. & 1--5 \\
\midrule
\textbf{Formality} & formality (decrease) & \textcolor{gray}{Yes, all good.} / Yup, all good. & 1--5 \\
\midrule
\multirow{ 2}{*}{\bf Complex.}  & reply length (shorter) & \textcolor{gray}{Yes, all good.} / Yes. & 1--5 \\
& reply length (longer) & \textcolor{gray}{Yes, all good.} / Yes, everything in the notes looks good.* & 1--5 \\
\midrule
\multirow{ 8}{*}{\bf Identity} & masculine marker & \textcolor{gray}{Yes, all good.} / Yes, all good man. & 1--3 \\
& feminine marker & \textcolor{gray}{Yes, all good.} / Yes, all good girl. & 1--3 \\
& pronoun (they/them) & \textcolor{gray}{Yes, all good.} / Yes, they did. All good. & 1--3 \\
& pronoun (he/him) & \textcolor{gray}{Yes, all good.} / Yes, he did. All good. & 1--3 \\
& pronoun (she/her) & \textcolor{gray}{Yes, all good.} / Yes, she did. All good. & 1--3 \\
& genderless relation & \textcolor{gray}{Yes, all good.} / Yes, all good friend.& 1--3 \\
& masc relation & \textcolor{gray}{Yes, all good.} / Yes, all good Dad.& 1--3 \\
& fem relation & \textcolor{gray}{Yes, all good.} / Yes, all good Mom. & 1--3 \\
& interests/habits & \textcolor{gray}{I'm sure it'll be fun.} / I'm sure you'll go to the beach. & 2--5$^{\dag\ddag}$ \\
\bottomrule
\end{tabular}
\caption{
Categories of observed system behaviors, informing the design of vignettes that we use to examine people's expectations of system behaviors. We use the first baseline reply as an anchor point for designing alternative reply options, which differ from the first reply along some subcategory.
$^\dag$CS2's assumptions around personal interests or behaviors are age related, e.g., mentions of driving. $^\ddag$CS3 includes assumptions with neutral or negative undertones. Examples marked with * operationalize differences in reply behaviors in response to \textit{I left you some notes. Is everything clear?}} 
\label{tbl:reply_categories}
\end{table}

Using the categories of system behaviors described above, we design crowdsourcing experiments to examine people's expectations of them (\textbf{RQ2}). These experiments are descriptive, encouraging participant subjectivity in order to capture a range of perspectives \cite{rottger-etal-2022-two}. We do not necessarily agree with all of the perspectives we surface (\S\ref{sec:limit}, \S\ref{sec:ethic}). However, these perspectives should inform considerations for how to navigate differing expectations when designing NLG systems.\looseness=-1

Each task instance shows a message containing an identity-related language feature and two reply options, which differ based on a category of system behaviors (Table~\ref{tbl:reply_categories}).
\footnote{As we are interested in people's expectations of system behaviors when we vary identity-related language features, our pilot experiments directly asked judges whether a reply was more usable for one message or another. However, instead of focusing on system behaviors (our goal), judges instead focused on the messages' perturbed language features, so we changed the task design. See Appendix~\ref{study_1:crowd} for details.}
One of the two reply options for each message is a baseline reply, which is a commonly generated reply with minimal modifications. The other reply operationalizes a subcategory of system behaviors; these are taken from actual systems' outputs or edited versions of the baseline reply.
Within each category of system behaviors, we investigate the same subcategories for CS4--5, and for CS1--3, with extra task instances involving personal interests or habits in CS2 and CS3, based on observed differences system behaviors (\S\ref{observed_section}).

We examine system behaviors in terms of \textbf{usability} \cite{robertson2021can} and \textbf{visibility}. For usability, we ask \textit{Which reply suggestion would you rather use as-is to reply to the message above?} with four options: the first reply, the second reply, both, or neither. Judges then select or specify reasons for why replies are unusable, which we use to validate the design of our reply options. 
Judges also write a reply they would send instead, and answer a binary question on visibility: whether the system should have blocked or shown the original unusable reply.

In addition to gathering judges' implicit reply preferences from curated message--reply vignettes, we gather explicit expectations of system behaviors by directly asking judges whether they generally believe that reply suggestions should adapt to a type of identity-related language feature in messages, and why. Other background questions focus on beliefs or lived experiences that may relate to judges' preferences: whether systems should infer gender from names (CS1), judges' familiarity with a name or country (CS1, CS3), and whether judges use an identity-related language feature (CS4, CS5).\footnote{While our task is descriptive, we use a few quality controls: we discard responses from judges who complete a HIT too quickly ($<$25 seconds), write nonsensical responses (e.g., keyboard smashing), misunderstand instructions, fail attention checks, or respond inconsistently to background questions.}
We collect three judgements for each task instance and target payments to match \$15 USD per hour. Across all five case studies, a total of 491 U.S.-based judges from Clickworker participated in our experiments. Full task instructions, reply options, and questions for CS1--5 are in Appendices~\ref{appdx:names}--\ref{appdx:style}. Throughout the rest of the paper, we highlight judges' remarks with quotations and italicized text.\looseness=-1

%% file: paper-files/sections/5.1_expected_behaviors.tex
\begin{figure}[t]
\centering
\includegraphics[trim= 0 7 0 32, clip, width=\columnwidth]{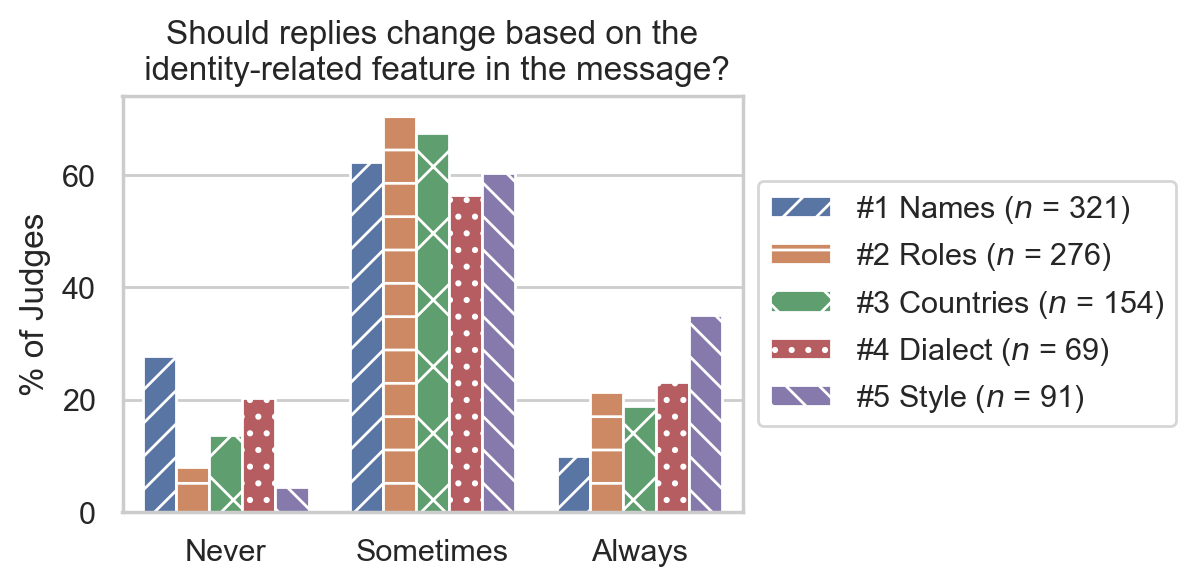} 
\caption{Distributions of judges' responses to whether they generally believe that reply suggestions should adapt to a
type of identity-related language feature.}
\label{fig:invariance}
\end{figure}

\begin{table*}[ht!]\footnotesize
\centering
\def\arraystretch{0.6}
\setlength{\tabcolsep}{0.5em}
\resizebox{\textwidth}{!}{
\begin{tabular}{@{}l c p{0.09\textwidth} p{0.38\textwidth} p{0.25\textwidth} p{0.5\textwidth}@{}} 
  & {\bf CS} & {\bf Category} & {\bf Message} &  {\bf Reply} & {\bf Description}\\ \toprule
 \multirow{40}{*}{\rotatebox[origin=c]{90}{\textbf{Adaptation}}} & 2 & Identity & It's been a good week. I won a soccer game. Best, \underline{Dad} & I'm so happy to hear that, \st{man!} \hlcolor{Dad!} & Social norms: ``\textit{Most would not address their father as if he were a peer.}'' The vocative \textit{man} can be too informal.\\ \cmidrule(lr){2-6}
 & 2 & Sentiment & It's been a good week. I won a soccer game. Best, \underline{Mom} & \textcolor{Gray}{\st{I'm so proud of you!}} \hlcolor{That is such good news!} & Social norms: ``\textit{Something you say that is appropriate to your kids may not be appropriate to your parents.}'' \\ \cmidrule(lr){2-6}
 & 3 & Formality & I'm from \underline{Eritrea} and I'm in town this week. What do you think about scheduling a meeting? & \textcolor{Gray}{\st{Sure, sounds cool.}} \hlcolor{Yes, I would be glad to meet then.} & Cultural differences: ``\textit{Someone with another native language may not understand the word cool in this context.}'' \\ \cmidrule(lr){2-6}
 & 3 & (Baseline) & What will I be doing in \underline{Eritrea} while visiting family? & \textcolor{Gray}{\st{I'm sure it'll be fun.}} \hlcolor{I suggest you go on a Safari.} & Feature-specific information: A reply suggestion system should know ``\textit{the details of every country or city in the world}''\\ \cmidrule(lr){2-6}
 & 5 & (Baseline) & Wha\underline{zzzzz UUUUUPPPPP!!!!} \underline{h}ow is everything in \underline{s}outh \underline{f}lorida? & \textcolor{Gray}{\st{Everything is good.}} \hlcolor{YO YO YO! AWESOME in South Florida!!!} & Accommodation: the judge indicated sentiment as the issue, and their adjusted reply includes similar stylistic features.\\ \cmidrule(lr){2-6}
 & 5 & (Baseline) & Call me. I forgot which meeting I should moderate. \underline{Helllllpppp.} & \st{Ok, will do!} \hlcolor{No worries, I'll call you asap!} & Accommodation: ``\textit{The sender's elongated hellllp makes the receiver understand the urgency of the matter.}''\\ \cmidrule(lr){2-6}
 & 4 & (Baseline) & I \underline{ain't} taking \underline{no} bus to come meet you. You better have a car. & \st{Ok, sounds good.} \hlcolor{No big, I gotta car} & Accommodation: this judge, who is not an AAE speaker, marked formality as the original reply's issue.\\ \cmidrule(lr){2-6}
 & 1 & Identity & I sent a follow up email last night. Did you get it? Best, \underline{Tony} & Yes, \st{thanks man.}  \hlcolor{thank you Tony} & Minimized assumptions: judges may avoid some assumptions, but allow replies re-stating references from the message.\\ \cmidrule(lr){2-6}
 & 2 & Identity & \underline{Mommy} wants to order dinner. Do you have any suggestions? & Yes, I'll give \textcolor{Gray}{\st{him}} \hlcolor{her} some ideas. & Minimized assumptions: judges may find parental terms less ambiguously gendered than names.\\ \midrule
 \multirow{12}{*}{\rotatebox[origin=c]{90}{\textbf{Invariance}}} & 1 & Identity & \underline{Kalen} sent a follow up email last night. Did you get it? & Yes, \textcolor{Gray}{\st{I'll thank them.}} \hlcolor{thanks.} & Prescriptivism: ``\textit{It does not sound like proper grammar.}'' Relatedly, perceived coherence is a common preference. \\ \cmidrule(lr){2-6}
 & 5 & Complexity & \underline{h}ey, what are you up to this weekend? & \st{No plans.} \hlcolor{I'm not sure yet.} & Prescriptivism: judges do not always accommodate messages' stylistic features in their adjusted replies. \\ \cmidrule(lr){2-6}
 & 1 & Identity & It's been a good week. \underline{Kalen} got promoted. & I'm so happy for \textcolor{Gray}{\st{him!}} \hlcolor{them!} & Avoid false assumptions: judges may replace \textit{he}/\textit{him} or \textit{she}/\textit{her} with \textit{they}/\textit{them} or no references.\\ \cmidrule(lr){2-6}
 & 3 & Identity & I'm from \underline{Afghanistan} and I'm in town this week. What do you think about scheduling a meeting? & Sure, let's meet at a \textcolor{Gray}{\st{bar.}} \hlcolor{nearby place} & Avoid false assumptions: ``\textit{Better not to assume anything... better not to assume someone is a drinker}'' \\ \bottomrule
\end{tabular}
}
\caption{Illustrative examples of judges' expectations of system behaviors in response to identity-related \underline{language features}. Each row pertains to one judge's response and explanation, if any, in quotes. Strikethrough text is not preferred, \textcolor{Gray}{text} the judge would rather not see is gray, and written \hlcolor{adjustments} are highlighted.} 
\label{tbl:expectations}
\end{table*}

\subsection{Mapping the Landscape of Expectations}
\label{judge_expect}

Figure~\ref{fig:invariance} summarizes judges' explicit expectations around whether replies should be invariant or adapt to identity-related language features.
Distributions of expectations vary for different types of features: Judges are more likely to favor adapting to style than to names. Dialect is a more polarizing case, where similar percentages of judges favor ``Never'' and ``Always'' and self-identified AAE speakers ($N=14$, CS4) are 21.6\% more likely to favor ``Sometimes'' or ``Always.'' This suggests that judges may not see invariance as a problem if they personally do not have a need for adaptation. In contrast, judges in CS5 who use any of the features in their own writing ($N=41$) are 26.8\% less likely to favor always adapting to style. 
Expectations also differ based on beliefs around the acceptability of making identity-related assumptions from a type of language feature. In CS1, judges who believe systems should never infer gender from names are 7.6 times more likely to respond ``Never'' to adaptation.\looseness=-1

All judges provide written, free-text explanations for their views. We summarize the major themes, which we obtain using iterative inductive coding of these explanations. First, we use open and axial coding to create thematic categories during an initial pass over all explanations; we then connect related themes and recode the explanations using the finalized categories for consistency. Where possible, we relate these explicit expectations to judges' implicit reply preferences, and provide illustrative examples in Table~\ref{tbl:expectations}. More detailed results for all five case~studies can be found in Appendices~\ref{appdx:names}--\ref{appdx:style}.\looseness=-1

\subsubsection{Adaptation} 

Broadly speaking, judges think adaptation can make replies 
more \textit{realistic}, \textit{natural}, \textit{authentic}, or \textit{genuine}, as ``\textit{[t]here's no one-size-fits-all}.''  
Reasons for adaptation include consideration of social norms, sensitivity to cultural differences, and awareness of feature-specific information. Judges also share potential strategies for adaptation, including facilitating linguistic accommodation, minimizing assumptions, and user-level adaptation. 


\textbf{Suggestions should follow social norms cued by features.}
In CS1--2, references can indicate the level of familiarity between people, like being on a ``\textit{first-name basis}'' with someone, or whether a situation is professional (\textit{doctor}, \textit{Mr.}) or casual (\textit{Mom}). References can thus evoke different levels of formality. Though judges believe systems should adhere to social norms, they have diverging beliefs about what those norms are. For example, ``\textit{when its the father [b]eing too informal might be a negative thing}'' is at odds with ``\textit{you can be less formal and use slang with family member.}'' While some judges advocate for more warmth within families, one judge says ``\textit{you can be short and to the point with family members or friends.}'' 
Informal replies are most often usable for \textit{Daddy}, \textit{Mom}, and \textit{Mommy}, while shorter replies are more often usable for parental terms than for names.\looseness=-1 


\textbf{Suggestions should be sensitive to cultural differences} 
and avoid unintended offense as ``\textit{certain phrases or customs that are acceptable in one country may be considered rude or inappropriate in another}'' (CS3). Judges note 
that different cultures may have different formality norms. Though judges value cultural sensitivity, their preferences are shaped by their awareness of cultural differences. For example, though replies mentioning \textit{drugs} are widely deemed inappropriate across countries, preferences around replies suggesting to \textit{meet at a bar} are highly varied without necessarily aligning with countries' cultural views on alcohol. 
Judges also suggest accounting for potential language barriers, e.g., by avoiding niche informal language or overloaded words 
(Table~\ref{tbl:expectations}).\looseness=-1



\textbf{Incorporating feature-specific information can make suggestions more helpful and appropriate.} In CS1, names ``\textit{could give clue as to [people's] race and gender,}'' and systems should 
avoid suggesting replies that ``\textit{could be inappropriate to certain races}.'' In CS3, suggestions could ``\textit{talk about things to do in certain countries}'' or adapt to time zones, events, and weather, e.g., ``\textit{get ready for the cold}.'' Some judges prefer replies suggesting activities (e.g., beach, hiking, museum) over more generic ones, and judge-adjusted replies offer other possibilities as well, including multiple mentioning pyramids in Egypt. One judge points out that wishing someone \textit{a fun trip} is more appropriate for tourist destinations while wishing someone \textit{a safe trip} is better for a country at war. However, in practice, judges rarely identified issues with the usability of intensely positive replies in response to travel (e.g., \textit{great trip!} or \textit{it'll be fun!}), even though the mentioned countries have varied associations with recent conflict.\looseness=-1


\textbf{Suggestions should help people attune or accommodate their language to each other},
such as converging on language style or word choice  \cite{giles_coupland_coupland_1991,danescu-niculescu-mizil-lee-2011-chameleons}. This theme 
is most common in CS5, where features can alter messages' affect, including their tone. For example, expressive elongation can make a message seem more ``\textit{young and hip and fun}'' or it can signal urgency (Table~\ref{tbl:expectations}). In addition, informal replies are deemed unusable in only 9.5\% of instances involving more casual messages, compared to 28.6\% of less casual ones. In CS4, when judges adjust replies to ``\textit{match},'' they sometimes attempt to write text that is more AAE-like (Table~\ref{tbl:expectations}).\looseness=-1


\textbf{Suggestions should minimize assumptions}, and can reuse references mentioned in a message (CS1--3). That is, a reply can contain \textit{Tony} if the message also uses this term. Judges emphasize consistency with user-established information, such as reusing pronouns previously assigned by the sender. Inferences can be made if they are considered sufficiently direct; judges vary in their beliefs around the extent to which replies should adapt to identity-related language features. For example, some judges believe \textit{Dad} is semantically (``\textit{distinctly}'') gendered and thus allows for \textit{he/him} pronouns, while names are more ambiguous.\looseness=-1


\textbf{Adaptation could occur at the user level}, since ``\textit{I choose options that sound like something I would say.}'' Judges suggest that systems could learn a user's interests, activities, or speech habits, or they could provide controllable identity-related settings. One judge suggests reply options could include ``\textit{a pull-down menu to choose him/her/them},'' which relates to a broader theme of how NLG systems could prioritize user agency in their design \cite{dudy-etal-2021-refocusing, robertson2021can}. In CS1, judges suggest that systems could recognize the names of a user's recurring contacts, and tailor replies based on prior conversations.\looseness=-1 


\textbf{Different types of features and other content should be considered together} when determining when and how adaptation should occur. However, this raises the question of how different types of features should be prioritized.  While judges in CS5 mention considering relative social roles, the reverse occurs in CS1--2, with some judges insisting that a message's style is more important.\looseness=-1 


\textbf{A lack of suggestions is not always undesirable.} Though a lack of reply suggestions can contribute to erasure \cite{schlesinger2018let}, it can also be perceived as a positive outcome in some contexts. Some judges do not want suggestions in casual situations, where the system may perceived be a ``\textit{nuisance}'' that prevents them from flexibly expressing themselves. Judges sometimes prefer no service to unusable service. For example, judges in CS1 wish to block replies that assume parental relationships in 80.7\% of unusable task instances.\looseness=-1


\subsubsection{Invariance} 

Invariance assumes the existence of general-purpose, ``\textit{default},'' ``\textit{neutral},'' or ``\textit{basic},'' suitable for all language features. Judges share several reasons for invariant system behaviors, ranging from prescriptivism to wariness of false assumptions.\looseness=-1

\textbf{Some judges take a prescriptive view}, wanting suggestions to be ``\textit{grammatically correct},'' using ``\textit{real}'' words and standard spelling, as 
``\textit{a more format (sic) and correct writing style is probably safer and more universal}.'' Correctness varies across language varieties. In the U.S., correctness may mean following style manuals and using GAE, promoted by predominantly white perspectives \cite{baron_2002_email, flores_2015_approp}.\looseness=-1


\textbf{Some judges think adaptation is unnecessary}, especially when an identity-related language feature (CS1--3) is not the focus of the message. For example, shorter replies that do not restate a name are sufficient. Generally, ``\textit{if someone has something additional to add, they can type it themselves.}'' Some judges also note that adaptation could increase cognitive load, as it may require people to check replies containing identity-related language features before sending. Favoring adaptation depends on whether judges expect it to lead to usable suggestions. One judge says that countries like Italy could have specific reply suggestions, but countries with a ``\textit{darker history}'' should not.\looseness=-1 


\textbf{Some judges believe adaptation is too difficult or complex} (CS2--5), so invariance is the best option: ``\textit{I don't think AI systems are advanced enough for this to work properly}.'' Still, a CS5 judge admits, ``\textit{it'd be pretty useful if it COULD pull it off.}'' Judges' beliefs around system behaviors are therefore affected by their perceptions of what systems can and cannot do.\looseness=-1



\textbf{Adaptation risks false assumptions, overgeneralization, and stereotypes}. Judges note many cases where identity-related language features are more ambiguous than expected, with one judge emphasizing ``\textit{DON'T ASSUME ANYTHING.}'' In CS1, the ethnic origin of a name ``\textit{does not mean that person grew up with that ethnic background.}'' In CS2, 
\textit{Daddy} can refer to a romantic partner or a father, and a parent--child relationship ``\textit{could be an estranged}'' one, making it difficult for parental roles to be mapped onto parental terms. Multiple judges indicate names are too vague to make assumptions, and ``\textit{commonly used gender pronouns may not always match how an individual wants to be identified}.'' In CS3, judges think that being from a country is not indicative of one's feelings of belonging to it or why someone is traveling, and suggestions of interests or activities should be avoided: ``\textit{you don't know what they are like, what they like to do, etc.}'' 
In CS4--5, linguistic accommodation can risk reply suggestions that include dialectal or stylistic features the user would never use, and in CS4, ``\textit{some people may find a non local (sic) entity speaking in dialect as offensive.}'' Indeed, nearly all judge-written replies in CS4 to AAE messages do not contain AAE or AAE-imitating features.\looseness=-1

Judges' reply preferences demonstrate how beliefs around assumptions involving identity-related language features can vary. For example, though 39.3\% of judges in CS1 think gender should never be inferred from names, others' reply preferences assume gender (Appendix~\ref{appdx:names_results}). In CS2, stereotypical pronouns for \textit{Michael} and \textit{Jennifer} are preferred at similar rates (41.1\%) to those for parental roles (43.9\%), contrary to some judges' stated belief that names are more ambiguously gendered.\looseness=-1


\textbf{Adaptation can cause discomfort and confusion}, even with supposedly valid replies. Suggestions that retain personal information can be ``\textit{creepy}'' or an ``\textit{invasion of privacy},'' especially if characteristics are correctly inferred based on indirect information. Adaptation can also confuse people who cannot discern why replies differ.\looseness=-1 


%% file: paper-files/sections/6_conclusion.tex
Through five case studies, in which we perturb different types of identity-related language features, we categorize a range of observed differences in NLG system behaviors and examine people's expectations around invariance and adaptation. 
People want systems to behave appropriately, but they diverge on what this entails and what assumptions systems should make. What some people view as a sociocultural norm, others may recognize as a stereotype, and some preferences, e.g., name-based gender inferences, conflict with current trends in fairness research \cite{lockhart2023name}. 
Accounting for people's lived experiences can help determine how we should translate their expectations of system behaviors into concrete recommendations for system design. Indeed, even our judges suggest drawing on participatory design methods \cite{muller1993participatory}, 
such as encouraging system developers to ``\textit{consult native speakers of the dialect}.''\looseness=-1



Our case studies focus on email reply as an illustrative downstream application, which allows us to surface expectations of NLG system behaviors within a specific context. For example, some judges in \S\ref{judge_expect} emphasize preserving user agency. Still, our findings also speak to other tasks or applications by questioning commonly held assumptions around how to specify desirable or ``good'' NLG model or system behaviors---of
which specifying ``fair'' behaviors is one example. Due to its simplicity, {\em invariance} may be an ``easy'' solution, where failing to exhibit the same system behaviors for different social groups is seen as unfair. 
{\em Adaptation}, where system behaviors should instead
vary across social groups, is a more open-ended, yet underexamined, challenge. 
When evaluating NLG systems, it is important to consider and discuss the implications of these assumptions. For example, as we show in \S\ref{judge_expect}, it is not always the case that the sentiment of system outputs should be invariant to identity-related language features in system inputs \cite{groenwold-etal-2020-investigating, sheng-etal-2021-societal}. Our findings open a path forward for more careful examination of both assumptions.\looseness=-1 

%% file: paper-files/sections/7-8_limitations_ethics.tex

\section{Limitations} \label{sec:limit}


\para{Limitations of using language features.} Our study follows the existing paradigm of operationalizing identity using only language features.
However, this paradigm involves many caveats discussed in prior work \citep[e.g.,][]{blodgett-etal-2021-stereotyping,goldfarb-tarrant-etal-2023-prompt}. 
For example, markers of majority groups, e.g., whiteness in U.S. contexts, are rarely explicitly stated in text \cite{mcdermott2022sociology}; official names of countries (CS3) may be complicated by political and diplomatic factors; and linguistic variables (CS4-5) can be linked to social identity with varying affective connotations and salience levels \cite{labov1972sociolinguistic,silverstein2003indexical, eckert2017phonetics}. Thus, our findings are limited to those we can surface with the subset of identity-related language features examined in each case study.\looseness=-1

\para{Limitations of our vignette-based design.}
In each task instance, we design each pair of reply options to operationalize differences based on a category of system behaviors (\S\ref{observed_section}).
However, we focus on text-only message--reply pairs in dyads and perturb individual language features in isolation, thus limiting ecological validity. 
We observe a few patterns in judges' responses that point to possible ecological validity issues (\S\ref{expected}).
To verify the design of each message--reply pair, we examine the reasons judges provide when they mark the second reply as not usable. Indeed, the provided reasons usually match the categories for which the pair was designed, but there are also cases where the distinctions between categories are not as clear cut. 
For instance, sentiment, affect, and text complexity can be conflated with formality, where warmer, shorter, and more intensely positive replies can be perceived as too informal (CS1, CS5). This is unsurprising since these broad characteristics are interconnected, with overlapping boundaries (\S\ref{observed_section}). The use of \textit{man} as a vocative is also perceived as both too informal and an inappropriate gender assumption (CS1--2), and some negated replies are perceived as incoherent (CS3--4). Stereotype-violating gender inferences (CS2--3) and the use of \textit{they} as a singular pronoun (CS1--2) may be perceived by some judges as incoherent, the latter echoing research on polarized views around nonbinary pronouns \cite{hekanaho2022thematic}. Thus, language differences are layered and tricky to isolate, as a single word can change multiple characteristics at once. \looseness=-1


\para{Limitations of judges' perspectives.} We use English-speaking, U.S.-based judges from Clickworker. To preserve privacy, we minimize the collection of demographic information from judges \cite{huang-etal-2023-incorporating}. 
Judges' expectations may not be reflective of the expectations of other populations or actual users of NLG systems, and
their perspectives are limited by their lived experiences. For example, one CS3 judge admits, ``\textit{I don't know much about Serbia but I think it's cold there.}''\footnote{The judge may have thought of Siberia, a region in Russia with cold winters; Serbia has a more subtropical climate.\looseness=-1} 

\section{Ethical Considerations}\label{sec:ethic}

While our work is IRB approved, we want to foreground several ethical considerations.
First, our work could be seen as suggesting that NLG systems {\em should} be used in applications involving interpersonal communications. However, prior work encourages reconsidering assumptions around whether some systems should be deployed at all \cite{barocas2020not, raji2022fallacy}.\looseness=-1


We also acknowledge that all names for dialects in CS4 necessarily encode sociopolitical commitments and are contested.
AAE consists of dialects that have also been given other labels by linguists and speakers over time, e.g., Ebonics, Black English, and African American Language. Similarly, GAE has also been given different labels by researchers, e.g., Mainstream American English and Standard American English.
While sociolinguists may use labels such as ``African American English'' to assert the dialects' systematicity and legitimacy (combating perceptions of ungrammaticality), such terms also take entire an ethnoracial group as their starting point and risk marking all group members' speech as non-normative \cite{king2020african}. Not all Americans of African descent are AAE speakers, and not all AAE speakers are African American. 

The AAE messages templates in CS4 are adapted from transcripts of Black AAE speakers (Appendix~\ref{appdx:dialect}).
We do not use synthetic examples, as AAE features have been stereotyped and appropriated in ways that erase their origins or disregard subtle aspects of how these features are actually used by AAE speakers---e.g., habitual \textit{be} being appropriated for non-habitual functions \cite{green2002african, sassy_queens2020, eberhardt2015first}.\looseness=-1


%% file: paper-files/sections/Appendix_A.tex
\section{Details for CS1 (Names)}\label{appdx:names}

\begin{table*}[th] \scriptsize
\centering
\begin{tabular}{>{\raggedright}p{4cm} p{11cm}} 
\toprule
  \textbf{Cluster label} & \textbf{Names} \\ \midrule
 South Asian & Syed, Nilesh, Abhishek, Vikram, Amit, Sangita, Ram, Parminder, Atul, Rama \\ \midrule
 East Asian (e.g. Korean, Japanese) & Cheuk, Jae, Wing, Sonny, Tan, Juanito, Yoon, San, Seong, Shin \\ \midrule
 Southeast Asian & Phan, Phuong, Quyen, Khang, Giang, Tuan, Kieu, Thang, Khoa, Vu \\ \midrule
 East Asian (e.g. Chinese) & Yong, Hao, Zhi, Shu, Yiu, Weiming, Zhong, Zhe, Mei, Zheng \\ \midrule
 White - European, masculine & Wilford, Deon, Robbie, Jeremy, Dixie, Clinton, Cameron, Harlan, Trent, Brad \\ \midrule
 White - Middle Eastern, masculine & Mitra, Rafi, Hany, Maha, Mansour, Hamid, Sami, Arash, Vahe, Sarı \\ \midrule
 White - European, feminine & Janey, Violet, Ramona, Annalisa, Abigail, Rita, Marlena, Natasha, Tena, Fern \\ \midrule
  White - European, masculine & Emanuel, Lucien, Marko, Pascal, Blaise, Panagiotis, Denis, Cristian, Angelika, Laurin \\ \midrule
  White - European, feminine & Cathi, Kandace, Stacey, Melodie, Kristyn, Tonja, Kathryn, Lyn, Wendie, Tressa \\ \midrule
  White - Central European, mix-gender & Alicja, Volodymyr, Darek, Wojciech, Nadezhda, Gordana, Veronika, Malgorzata, Bohdan, Grzegorz \\ \midrule
  Hispanic - masculine & Marcelo, Norberto, Flavio, Pascual, Gerardo, Fredy, Marcos, Ramiro, Amador, Efren \\ \midrule
  Hispanic - feminine &  Ernestina, Haydee, Ines, Yolanda, Guadalupe, Maritza, Noemi, Eliana, Arcelia, Leonor \\ \midrule
  Other - masculine & Eddy, Augustin, Dexter, Renato, Salim, Rico, Quincy, Linwood, Khalid, Rene \\ \midrule
  Other - feminine & Ester, Violeta, Aurelia, Milagros, Dalia, Salina, Annie, Lisette, Jacinta, Evette \\ \midrule
  Black - mixed gender w/ mostly masculine & Sylvester, Mable, Alfreda, Cornell, Tyrone, Darnell, Lula, Alphonso, Althea, Demetrius \\ \midrule
  Black - feminine & Lawanda, Earnestine, Marva, Lakisha, Latrice, Tanisha, Jamila, Keisha, Jermaine, Latoya \\ \bottomrule
\end{tabular}
\caption{Clusters with example names and descriptions corresponding to race, regional, and gender associations.}
\label{tbl:name_clusters}
\end{table*}

\begin{table*}[th] \scriptsize
\centering
\begin{tabular}{ p{6cm}  p{5.6cm}  p{2.6cm}} 
\toprule
\multicolumn{2}{c}{\textbf{Message template}}                                     &  \\ \cmidrule(r){1-2}
\textbf{First person (sender \& recipient)} & \textbf{Third party} & \textbf{Baseline reply} \\ \midrule
 I sent a follow up email last night. Did you get it?
 & \textsc{person} sent a follow up email last night. Did you get it? & Yes, thank you. \\ \midrule
 I left you some notes. Is everything clear?
 & \textsc{person} left you some notes. Is everything clear? & Yes, all good. \\ \midrule
 It's been a good week. I got promoted.
 & It's been a good week. \textsc{person} got promoted. & I'm so happy to hear that!\\ \midrule
 I got into an accident while on vacation. Ended up breaking both an arm and a leg.
 & \textsc{person} got into an accident while on vacation. Ended up breaking both an arm and a leg. & I'm sorry to hear that.\\ \midrule
 I am in town this week. What do you think about scheduling a meeting?
 & \textsc{person} is in town this week. What do you think about scheduling a meeting? & Sure, sounds good.\\  \bottomrule
\end{tabular}
\caption{Message templates used for CS1. The baseline reply is used for crowdsourcing, where judges compare this reply with a second reply that differs along some category of reply behavior, such as sentiment or formality.}
\label{tbl:name_messages}
\end{table*}

\begin{table} \scriptsize
\centering
\begin{tabular}{ p{1.2cm} p{5cm}} 
\toprule
  \textbf{Position} & \textbf{Example}  \\ \midrule
  Sender & It will be a long day. I'll bring snacks for everyone. Best, \textcolor{BrickRed}{\textbf{Jennifer}}\\\midrule
  Recipient & Hi \textcolor{BrickRed}{\textbf{Jennifer}},
It will be a long day. I'll bring snacks for everyone.\\\midrule
  Third party & It will be a long day. \textcolor{BrickRed}{\textbf{Jennifer}} will bring snacks for everyone. \\
  \bottomrule
\end{tabular}
\caption{For CS1--2, we place references to a person in three different positions in messages: the sender, the recipient, or a third party being mentioned.}
\label{tbl:name_directions}
\end{table}

\subsection{Messages}

\para{Feature selection.} To address RQ1, we obtain a sample of names that cover a range of ethnic and gender connotations. First, we obtain a potential pool of names from a dataset of first names used in mortgage applications, and each name is labeled with the percentage of individuals with that name who belong to six race and ethnicity categories \cite{tzioumis2018demographic}. These categories include Hispanic, non-Hispanic (NH) white, NH Black or African American, NH Asian or Native Hawaiian or Other Pacific Islander, NH American Indian or Alaska Native, and NH multi-racial. 

Next, we perform stratified sampling of names from clusters induced from this collection. Word embeddings for names can cluster based on shared sociodemographic associations \cite{romanov-etal-2019-whats}. For each name, we label it as associated with a race/ethnicity if at least 50\% of the people with that name in the dataset fall under that race/ethnicity. We then cluster names' fastText embeddings within racial/ethnic categories \cite{bojanowski-etal-2017-enriching}, choosing a number of clusters where groupings roughly correspond to different genders and regions (Table~\ref{tbl:name_clusters}). 

The descriptive labels for each cluster in  Table~\ref{tbl:name_clusters} describe potential sociodemographic connotations of names. To identify regional associations, we manually inspected a sample of around ten names from each cluster and their Wikipedia pages for information on origin and use, if present. To identify binary gender associations, we use U.S. birthname lists for gender and examine the proportion of names in each cluster tend to be majority ($>$75\%) feminine or masculine in these lists \cite{ssa_names}. Clusters for East and Southeast Asian names contain both masculine and feminine names, while other clusters tend to lean more heavily towards one gender. From each cluster, we sample at least 15 names to use in input messages for each system. 

\para{Message design.} We input names into a subset of message templates created by \citet{robertson2021can}, picking those that do not include third-person pronouns (Table~\ref{tbl:name_messages}). These two-sentence message templates are formatted to contain some context for the message, followed by a speech act common in emails, such as a question, notification, or request. We choose 5 message templates that cover different speech acts: a binary question about receiving an email, a binary question around clarity, a notification of a positive event, a notification of a negative event, and a request to schedule a meeting. 

Names can be mentioned in the greeting, main body, and closing of emails (Table~\ref{tbl:name_directions}). We leverage this structure to construct 3 versions of each message template as inputs into reply generation systems (RQ1). For senders, we append the the closing \textit{Best, [name]}, and for recipients, we prepend the greeting \textit{Hi, [name]}. For third party mentions, we replace first-person references in these message templates with names, modifying verb forms if needed. 

\subsection{Crowdsourcing design} \label{study_1:crowd}

\begin{figure}[th]
\centering
\includegraphics[width=\columnwidth]{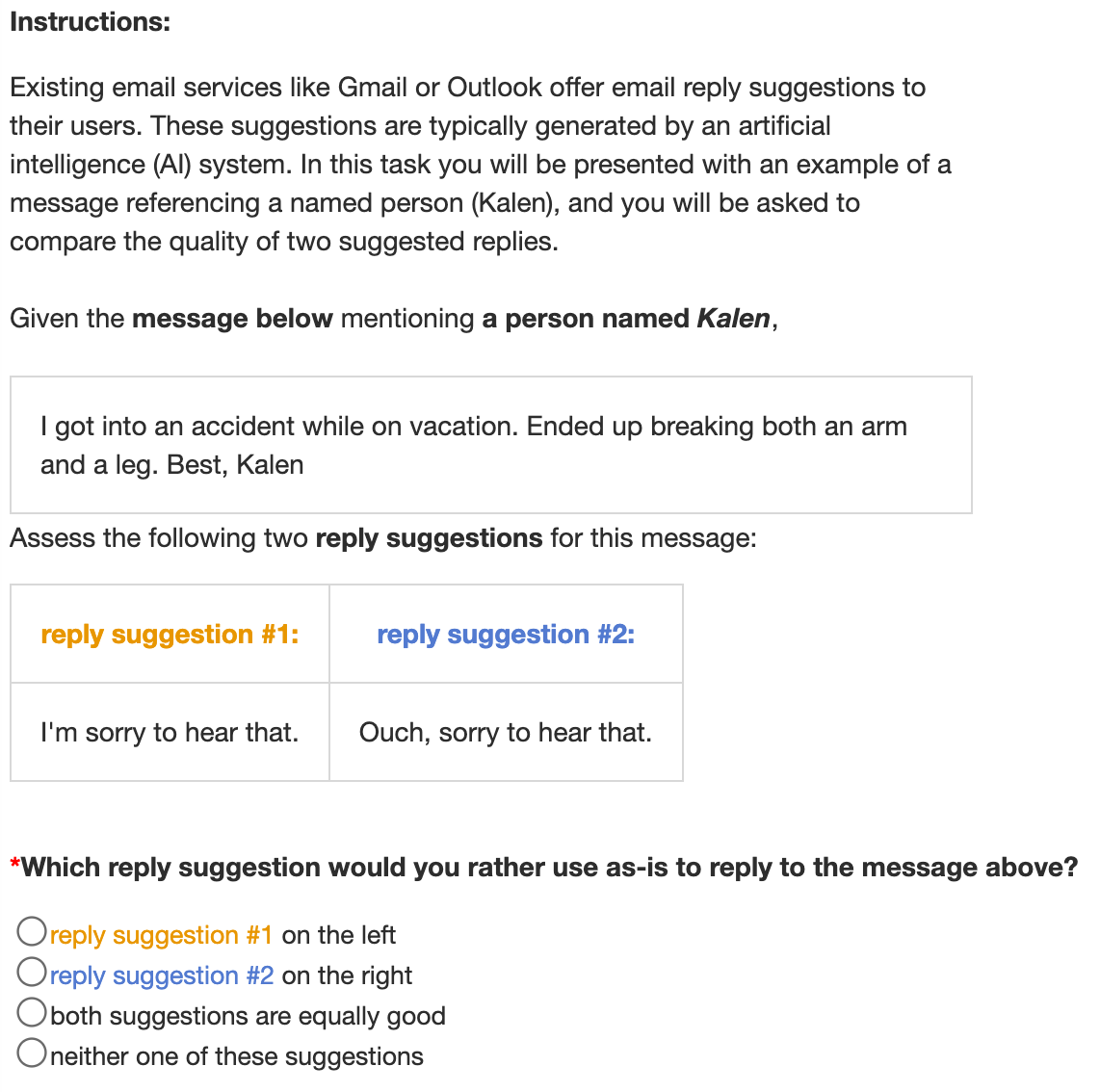}

\includegraphics[width=\columnwidth]{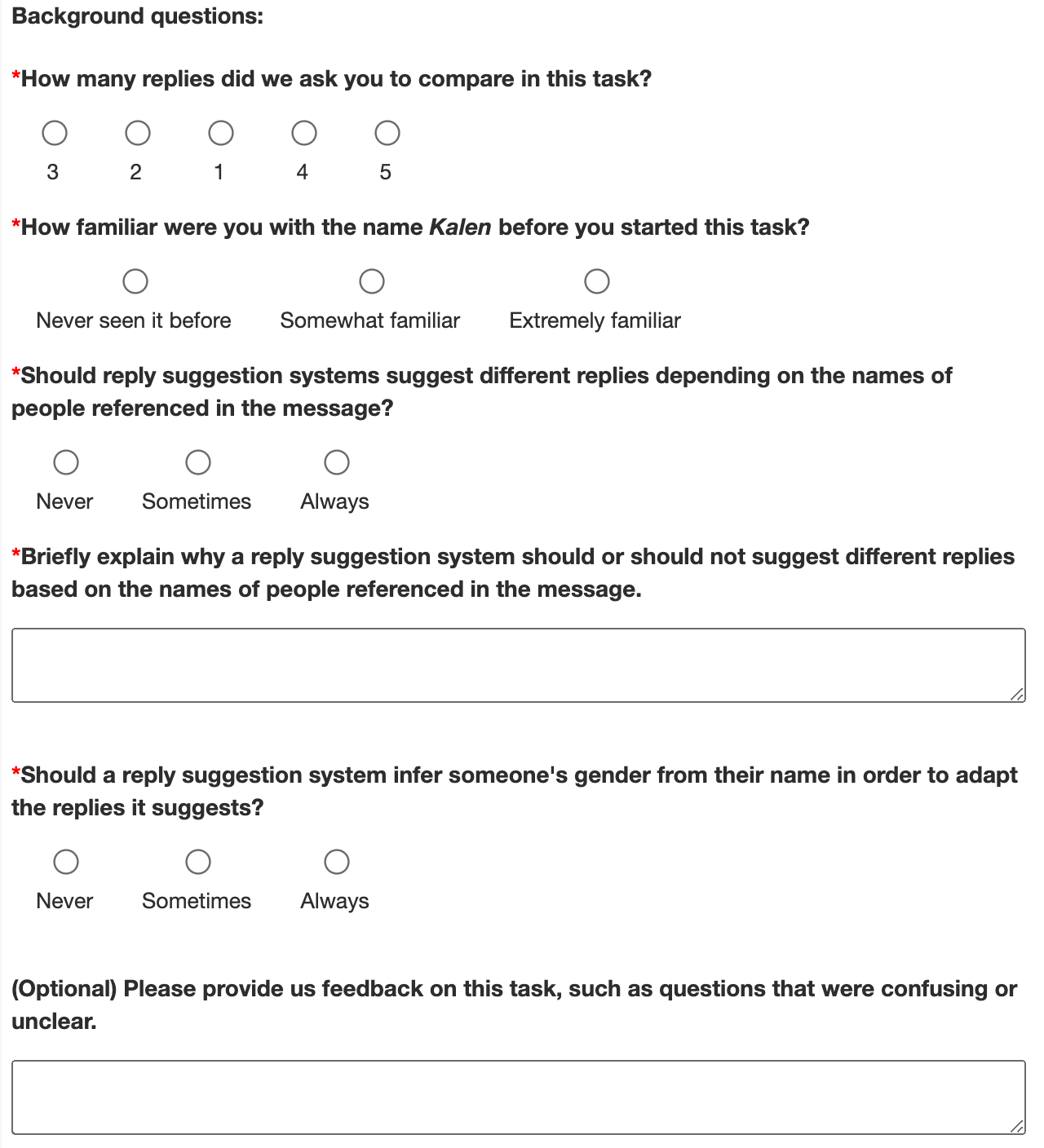}
\caption{Main body of task instructions and questions in CS1. Other case studies use a similar format.}
\label{fig:CS1_questions1}
\end{figure}

\begin{figure}[th]
\centering
\includegraphics[width=\columnwidth]{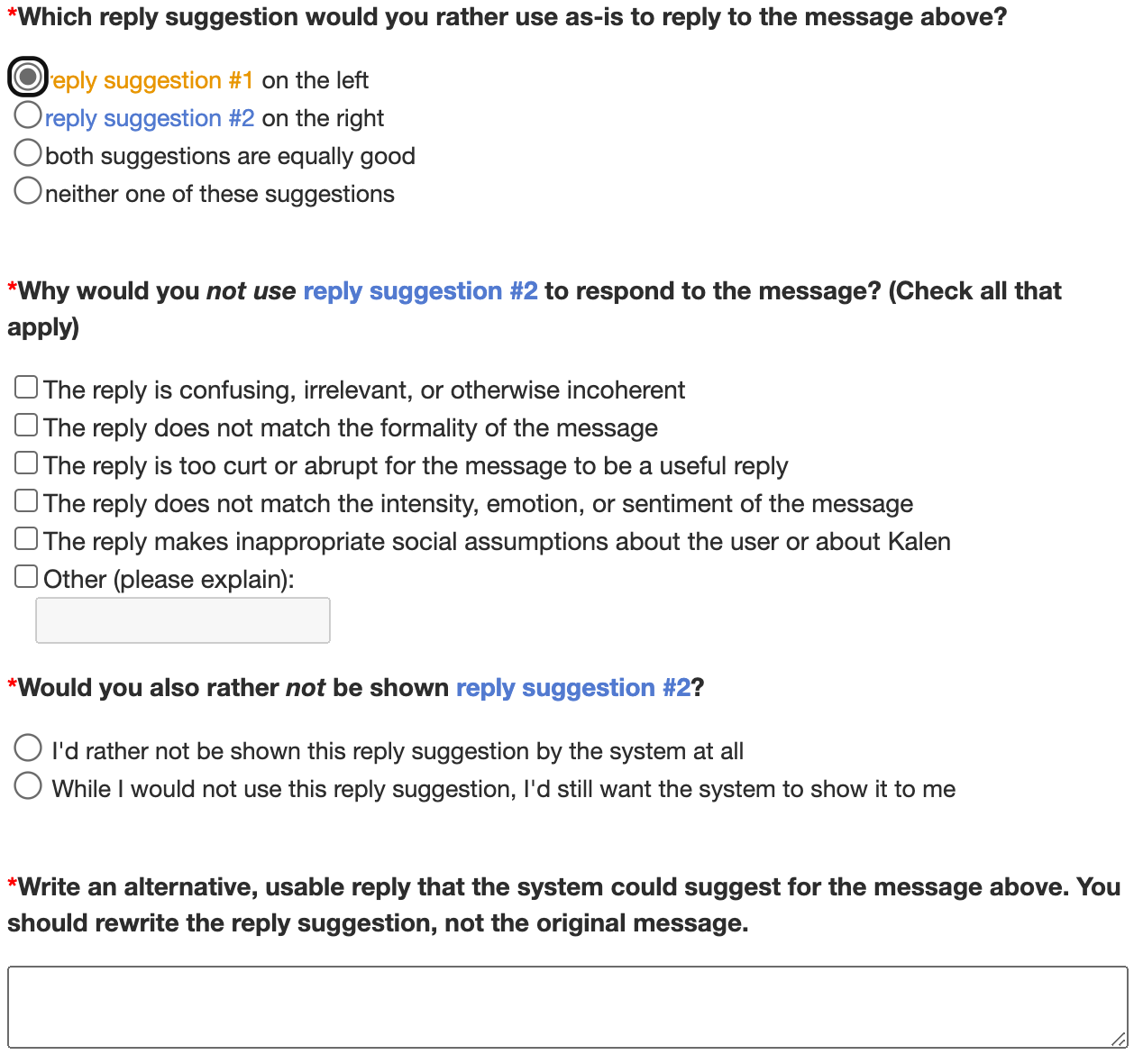}
\caption{Additional followup questions when at least one reply is deemed more usable. In this example, \textit{reply suggestion \#1} is selected, so followup questions target the usability of \textit{reply suggestion \#2}.}
\label{fig:CS1_questions3}
\end{figure}

We use CS1 pilot experiments to establish our crowdsourcing task design for all case studies. In these pilots, we ask judges to directly compare the usability of a reply given two messages containing different names, but this leads to some judges stating that a reply is less usable for a message because the message contains a ``\textit{bad}'' or unusual name. A similar phenomenon occurs when piloting this initial design with CS4, where some judges state that proposed replies are more usable for the GAE message because they believe the AAE message is ungrammatical. Thus, to de-emphasize preferences around the identity-related feature itself, we shift to the task design we describe in the main text, which examines implicit preference differences around reply behaviors.

The instructions and body of this task can be viewed in Figures~\ref{fig:CS1_questions1} and \ref{fig:CS1_questions3}. They are also written in the following text: 

\begin{displayquote}
    \textit{Existing email services like Gmail or Outlook offer email reply suggestions to their users. These suggestions are typically generated by an artificial intelligence (AI) system. In this task you will be presented with an example of a message referencing a named person (}\textsc{Name}\textit{) and you will be asked to compare the quality of two suggested replies.}

    \textit{Given the message below mentioning} \textsc{Name},

    \textsc{Message}

    \textit{Assess the following two reply suggestions for this message:}

    \textsc{Baseline Reply} $\|$ \textsc{Second Reply}

    \textit{Which reply suggestion would you rather use as-is to reply to the message above?} Single-choice options: \textit{reply suggestion \#1 on the left}; \textit{reply suggestion \#2 on the right}; \textit{both suggestions are equally good}; \textit{neither one of these suggestions}. 
\end{displayquote}

If \textit{reply suggestion \#2} or \textit{neither} is selected to the previous question, we show these followup questions: 

\begin{compactitem}
\item \textit{Why would you not use reply suggestion \#1 to respond to the message? (Check all that apply)}. Options: \textit{The reply is confusing, irrelevant, or otherwise incoherent}; \textit{The reply does not match the formality of the message}; \textit{The reply is too curt or too abrupt for the message to be a useful reply}; \textit{The reply does not match the intensity, emotion, or sentiment of the message}; \textit{The reply appears to make inappropriate social assumptions about the user or about \textsc{Name}}; \textit{Other (please explain)}. 
\item \textit{Would you also rather not be shown reply suggestion \#1?} Single-choice options: \textit{I'd rather not be shown this reply suggestion by the system at all}; \textit{While I would not use this reply suggestion, I'd still want the system to show it to me}.
\item \textit{Write an alternative, usable reply that the system could suggest for the message above. You should rewrite the reply suggestion, not the original message.} Free response box. 
\end{compactitem}

A similar set of followup questions is shown if reply \#1 or \textit{neither} is instead selected as more usable, except with \textit{reply suggestion \#2} mentioned instead of \textit{reply suggestion \#1}.

\begin{figure}[t]
\centering
\includegraphics[width=0.9\columnwidth]{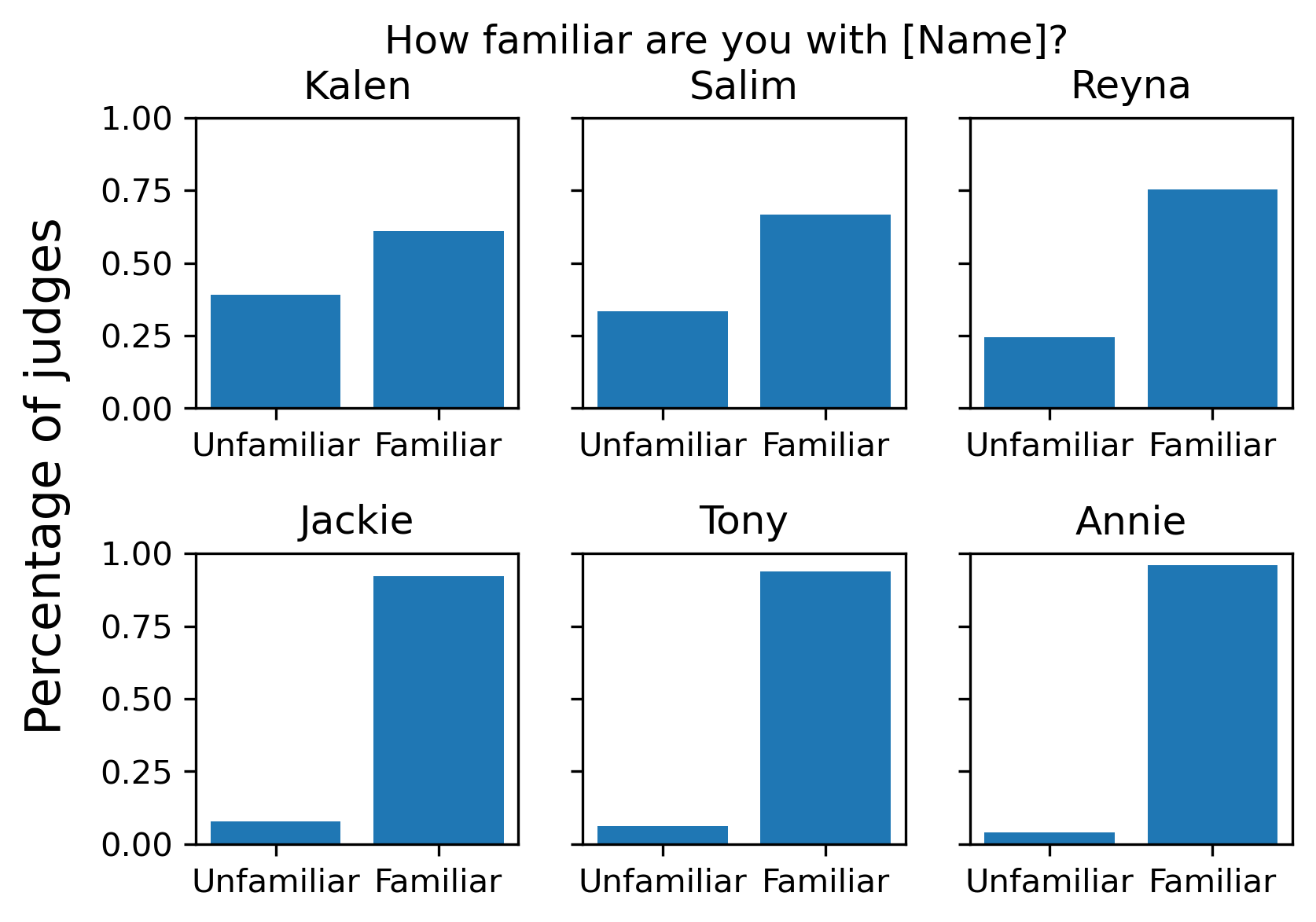}
\caption{The six names tested during the crowdsourcing phase of CS1 evoke different levels of familiarity among judges. The $x$-axis binarizes responses so that \textit{Unfamiliar} corresponds to responding \textit{Never seen it before}, while \textit{Familiar} corresponds to \textit{Somewhat} or \textit{Extremely familiar}.}
\label{fig:CS1_familiarity}
\end{figure}

\begin{figure}[t]
\centering
\includegraphics[width=0.9\columnwidth]{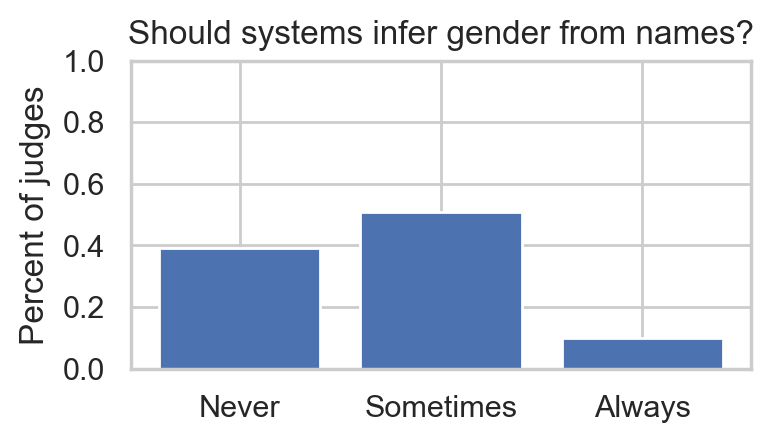}
\caption{Around 39.7\% of judges in CS1 believe that reply suggestion systems should never infer gender from names.}
\label{fig:CS1_gender}
\end{figure}

Background questions for CS1 include the following: 

\begin{compactitem} 
    \item \textit{How many replies did we ask you to compare in this task?} Single-choice options: 1, 2, 3, 4, 5 in randomized order. This is an attention check, where the correct answer is \textit{2}.
    \item \textit{How familiar were you with the name} \textsc{Name} \textit{before you started this task?} Single-choice options: Never seen it before, Somewhat familiar, Extremely familiar (Figure~\ref{fig:CS1_familiarity}).
    \item \textit{Should reply suggestion systems suggest different replies depending on the names of people referenced in the message?} Single-choice options: Never, Sometimes, Always (Figure~\ref{fig:invariance}).
    \item \textit{Briefly explain why a reply suggestion system should or should not suggest different replies based on the names of people referenced in the message.} Free response box.
    \item \textit{Should a reply suggestion system infer someone's gender from their name in order to adapt the replies it suggests?} Single-choice options: Never, Sometimes, Always (Figure~\ref{fig:CS1_gender}).
    \item \textit{(Optional) Please provide us feedback on this task, such as questions that were confusing or unclear.} Free response box.
\end{compactitem} 

The attention check and free response box around why a system should or should not adapt to names were added to the task after we collected 65\% of total judgements for this case study. The first addition was useful for more efficient filtering of spammers, and the latter was useful for addressing RQ2. The final task design for CS1 was then used as a basis for later case studies. 

\begin{figure}[t]
\centering
\includegraphics[trim= 0 2 0 4, clip, width=0.9\columnwidth]{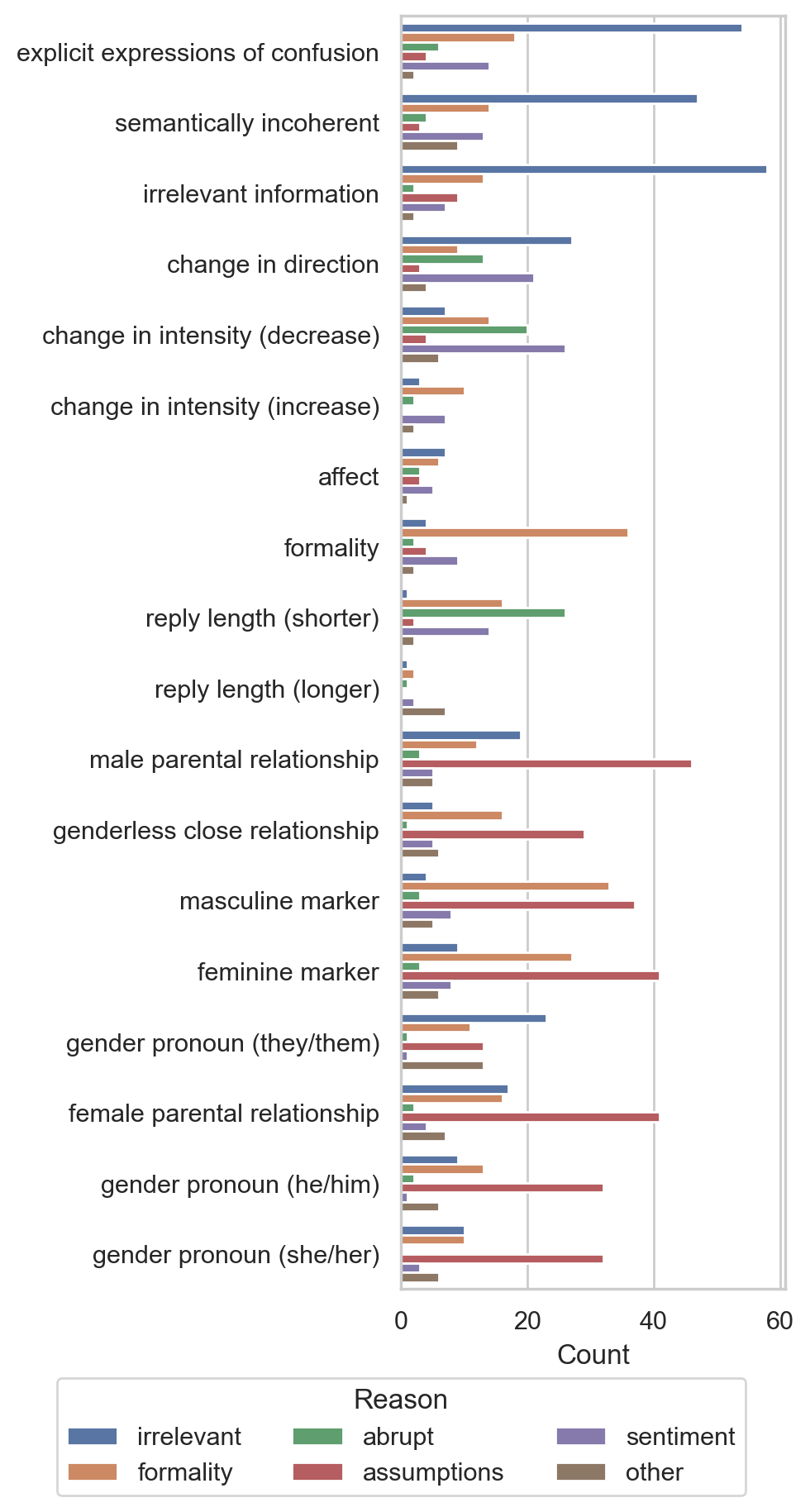}
\caption{Reasons judges marked the second reply as less usable or not usable in CS1. The second reply differs from the baseline reply option along the subcategory of reply behavior shown on the $y$-axis.}
\vspace{-5pt}
\label{fig:CS1_reasons}
\end{figure}

Occasionally judges would change their Likert responses to background questions across task examples. These judges' written responses were generally valid, so these changes may be cases where their opinion has changed after encountering additional task examples. Thus, we take the average Likert scale rating for each judge and background question, and round it to the nearest integer to represent a judge's overall rating. 

\subsection{Crowdsourcing results}\label{appdx:names_results}

\para{Reply pair validity.} Figure~\ref{fig:CS1_reasons} shows the frequency of various reasons being checked for unusable second replies modified from baseline replies. As discussed in \S\ref{expected}, we use this to examine the validity of how we operationalized reply behaviors. The reason second replies were unusable most often followed the intention of our design, with a few exceptions. Negative or warmer replies, and those that use \textit{they}/\textit{them} pronouns can be often perceived as incoherent. In addition, the masculine marker \textit{man} could be perceived as not just assumptious, but also too informal. 

\begin{figure}[t]
\centering
\includegraphics[width=0.48\columnwidth]{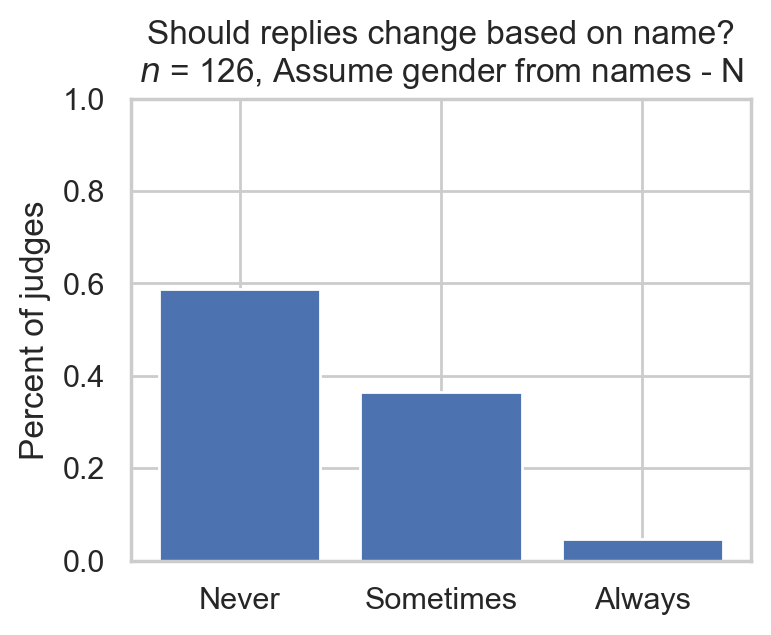}
\includegraphics[width=0.48\columnwidth]{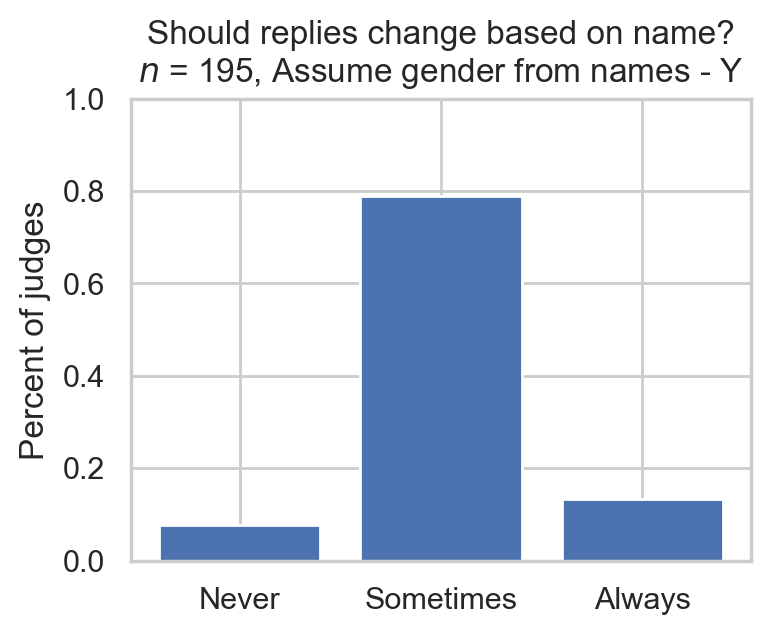}
\caption{Judges' beliefs around invariance and adaptation shift depending on whether they believe it is acceptable to infer gender from names. Here, ``N'' corresponds to \textit{Never} in response to the background question in Figure~\ref{fig:CS1_gender}, while ``Y'' corresponds to \textit{Sometimes} or \textit{Always}.}
\label{fig:CS1_background}
\end{figure}

\para{Responses to background questions.} To address RQ2, messages perturbed six names that reflect not only varying gender connotations, but were also likely to evoke different levels of familiarity among judges (Figure~\ref{fig:CS1_familiarity}). When it comes to systems inferring gender from names, judges' responses were mostly split between ``Never'' and ``Sometimes'' making these assumptions (Figure~\ref{fig:CS1_gender}). In addition, judges who believe gender should never be inferred from names are less likely to favor adaptation than invariance (Figure~\ref{fig:CS1_background}). 

\begin{figure}[t]
\centering
\includegraphics[trim= 0 4 0 4, clip, width=\columnwidth]{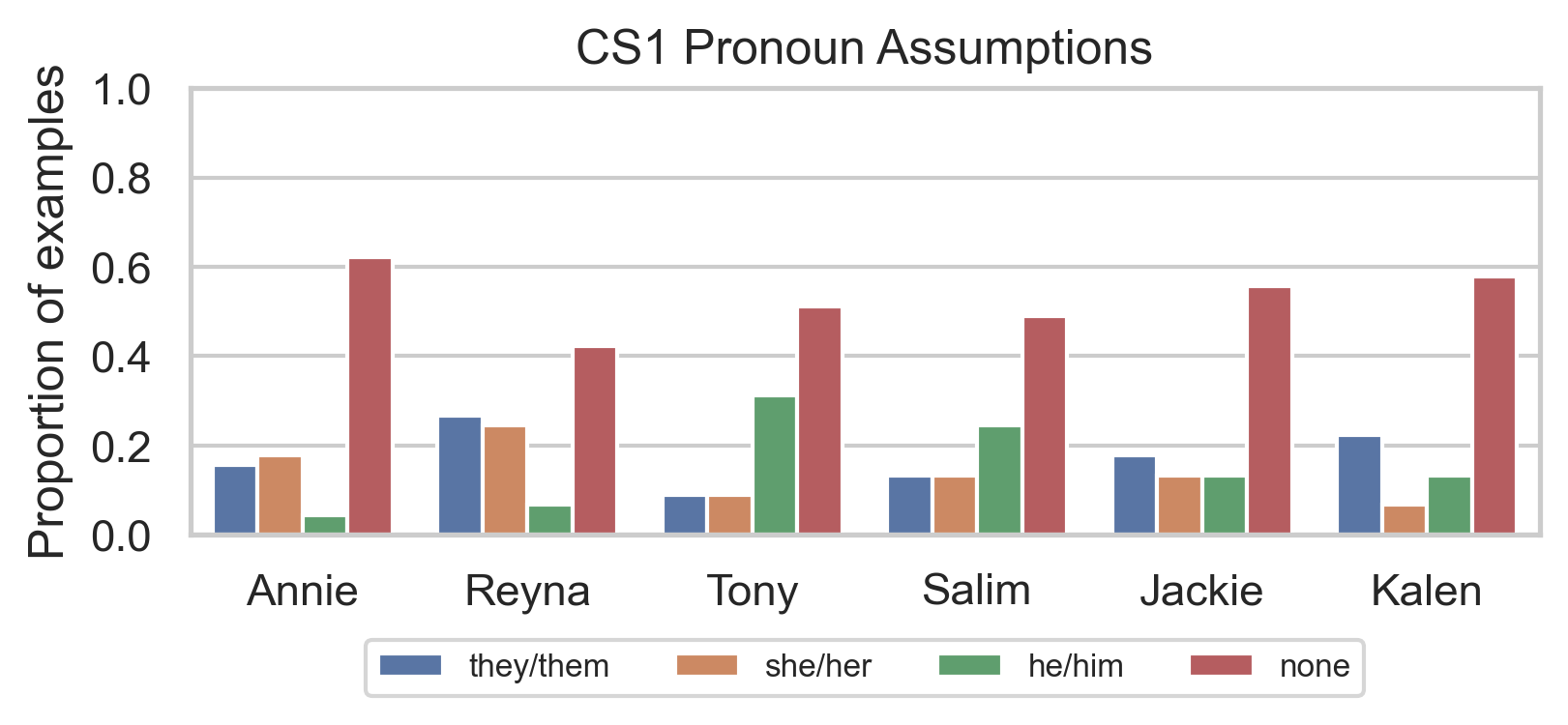}
\caption{Replies deemed equally usable or preferred compared to a pronoun-less baseline reply in CS1. These include judges' edited replies, e.g. \textit{Yes, I'll respond to \underline{him} soon.}}
\vspace{-8pt}
\label{fig:CS1_pronouns}
\end{figure}

\para{Aggregated reply preferences.} In \S\ref{judge_expect}, we use judges' free written responses to guide what subcategories to investigate further. Judges' written responses were especially verbose when reply options assumed gender, such as around pronouns. Though a substantial proportion of judges did not include any pronouns in their preferred or edited replies, others did, and sometimes in a stereotype-aligned manner, e.g. \textit{he/him} with \textit{Tony} (Figure~\ref{fig:CS1_pronouns}). Additional results juxtaposing stereotype-violating assumptions across CS1--2 can be found in \S\ref{appdx:roles_results}.

\begin{figure*}
\centering
\includegraphics[width=0.9\textwidth]{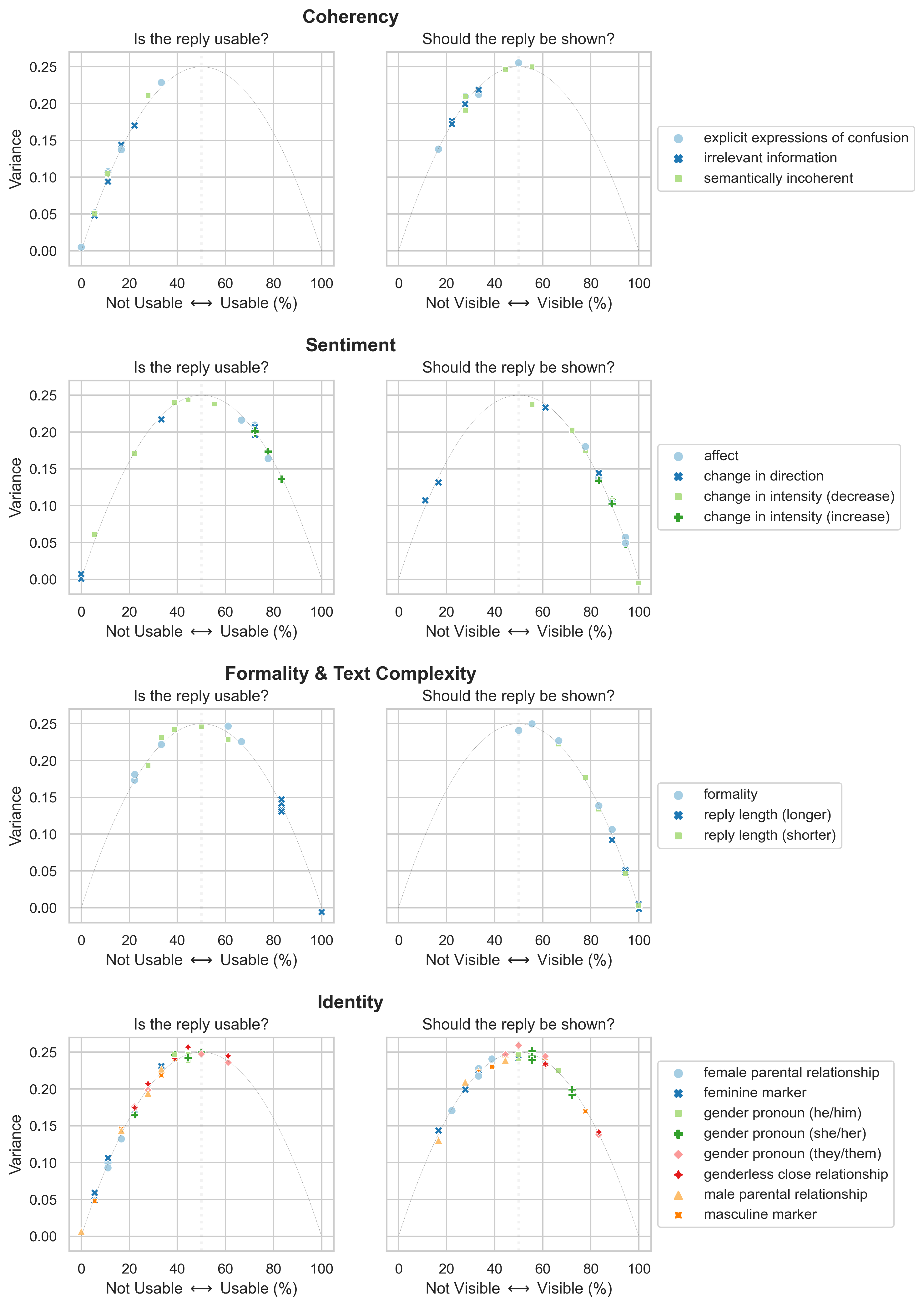}
\caption{In these plots, each point is a message template, and the probability ($x$-axis) that a second reply option is usable (left) and visible (right) is aggregated across different perturbations of names (CS1). When there is less consensus around a reply behavior, variance ($y$-axis) is high. A light vertical gray line emphasizes the highest possible variance, and jitter is added along the $y$-axis so that overlapping points are more visible.}
\label{fig:CS1_variance}
\end{figure*}

A bottom-up view of reply preferences also reveals additional insights. Figure~\ref{fig:CS1_variance} shows aggregated results around the visibility and usability of replies across names. Statistical variance has been used in prior work to measure annotator disagreement \cite{davani-etal-2022-dealing}, and higher and lower probabilities have lower variance. There are a few takeaways from this overview:

\begin{compactitem}
    \item As expected, incoherent replies are typically not usable, though explicit expressions of confusion, e.g. \textit{I'm not sure what you mean by this}, are not always recognized as unusable. 
    \item Most sentiment categories are usually usable and should be suggested, except for less intense replies and a few negative replies. A leaning towards more positive reply suggestions has also occurred in previous work observing smart-reply systems \cite{hohenstein_2018_ai}.
    \item Longer replies are usually more usable, while informal ones are less, and the latter case may be due to the topic of the messages we tested, as they tend to pertain to professional settings.
    \item Identity-related assumptions span a range of usability that is similar to that of incoherence. The use of the feminine marker \textit{girl} and \textit{Mom} are especially undesirable, while the assumption of different pronouns varies highly. Less gendered assumptions, e.g. \textit{they/them} and \textit{friend}, can be less preferred but still often allowed to be suggested. 
\end{compactitem}

%% file: paper-files/sections/Appendix_B.tex
\clearpage
\section{Details for CS2 (Parental Roles)}\label{appdx:roles}

\begin{table*}[th] \scriptsize
\centering
\begin{tabular}{ p{5cm}  p{5cm}  p{4cm}} 
\toprule
\multicolumn{2}{c}{\textbf{Message template}}                                     &  \\ \cmidrule(r){1-2}
\textbf{First person (sender \& recipient)} & \textbf{Third party} & \textbf{Baseline reply} \\ \midrule
 I'm leaving now. We'll be at the restaurant soon.
 & \textsc{person} is leaving now. We'll be at the restaurant soon. & Okay, sounds good.\\ \midrule
 I want to order dinner. Do you have any suggestions?
 & \textsc{person} wants to order dinner. Do you have any suggestions?  & Yes, I do.\\ \midrule
 It's been a good week. I won a soccer game.
 & It's been a good week. \textsc{person} won a soccer game.  & I'm so happy to hear that!\\ \midrule
 I want to get together and talk. When are you free?
 & \textsc{person} wants to get together and talk. When are you free? & Sure, I'm free now.\\ \midrule
 It will be a long day. I'll bring snacks for everyone.
 & It will be a long day. \textsc{person} will bring snacks for everyone. & Okay, thank you!\\ \bottomrule
\end{tabular}
\caption{Message templates used for CS2 (parental roles). The baseline reply is used to crowdsource preferences around a range of reply behaviors, such as those listed in Table~\ref{tbl:reply_categories}.}
\label{tbl:relative_messages}
\end{table*}

\subsection{Messages}

This case study's design parallels that of CS1. We crafted five two-sentence message templates inspired by those used by \citet{robertson2021can}, changing  workplace-related terms with ones that would be more likely to be used among family members (Table~\ref{tbl:relative_messages}). The terms for parental roles and names (\textit{Mommy}, \textit{Mom}, \textit{Jennifer}, \textit{Daddy}, \textit{Dad}, \textit{Michael}) were placed in sender, recipient, and third party positions in these message templates (Table~\ref{tbl:name_directions}). 

\subsection{Crowdsourcing design}

The instructions for this task is the following, where \textsc{Person} is a name or parental role: 

\begin{displayquote}
\textit{Existing email services like Gmail or Outlook offer email reply suggestions to their users. These suggestions are typically generated by an artificial intelligence (AI) system. In this task you will be presented with an example of a message referencing a family member or named individual, and you will be asked to compare the quality of two suggested replies.}

\textit{Given the message below mentioning} \textsc{Person},

\textsc{Message}

\textit{Assess the following two reply suggestions for this message:}

\textsc{Baseline Reply} $\|$ \textsc{Second Reply}
\end{displayquote}

After these instructions, the body of the task matches CS1. The background questions for this case study are the following: 

\begin{compactitem}
    \item \textit{How many replies did we ask you to compare in this task?} Single-choice options: 1, 2, 3, 4, 5 in randomized order. This is an attention check, where the correct answer is \textit{2}.
    \item \textit{Depending on their relationships with others, the same person may be referred to using different terms, such as their occupation (Doctor), their familial role (Mom, Mommy), or their own name (Jessica). Should reply suggestion systems suggest different replies based on how someone is referred to?} Single-choice options: Never, Sometimes, Always (Figure~\ref{fig:invariance}). 
    \item \textit{Briefly explain why a reply suggestion system should or should not suggest different replies based on how someone is referred to in the message.} Free response box.
    \item \textit{(Optional) Please provide us feedback on this task, such as questions that were confusing or unclear.} Free response box.
\end{compactitem}

\subsection{Crowdsourcing results}\label{appdx:roles_results}

\begin{figure}[t]
\centering
\includegraphics[trim= 0 2 0 4, clip, width=0.9\columnwidth]{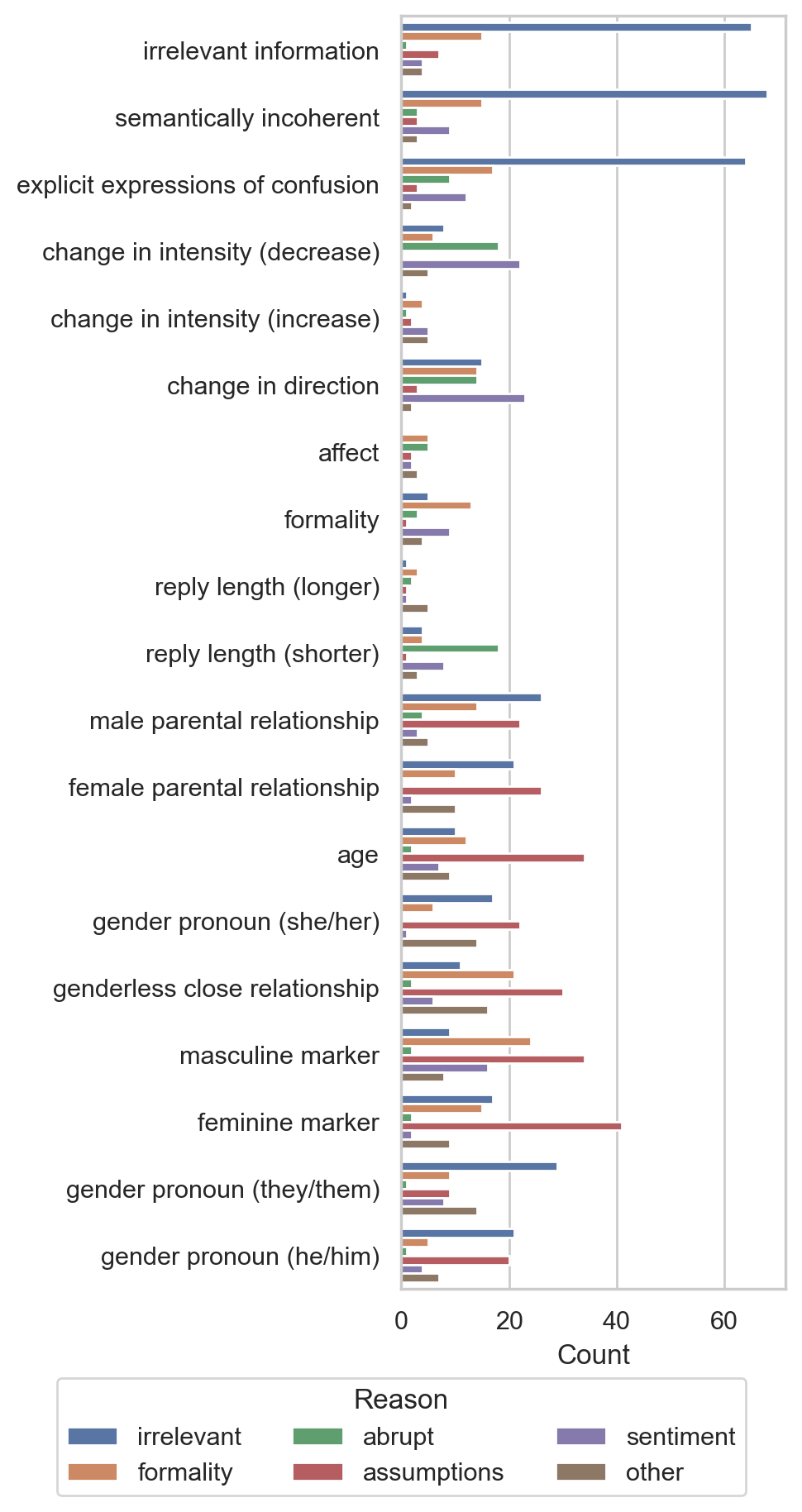}
\caption{Reasons judges marked the second reply as less usable or not usable in CS2. The second reply differs from the baseline reply option along the subcategory of reply behavior shown on the $y$-axis.}
\vspace{-5pt}
\label{fig:CS2_reasons}
\end{figure}

\begin{figure}[t]
\centering
\includegraphics[trim= 0 4 0 4, clip, width=\columnwidth]{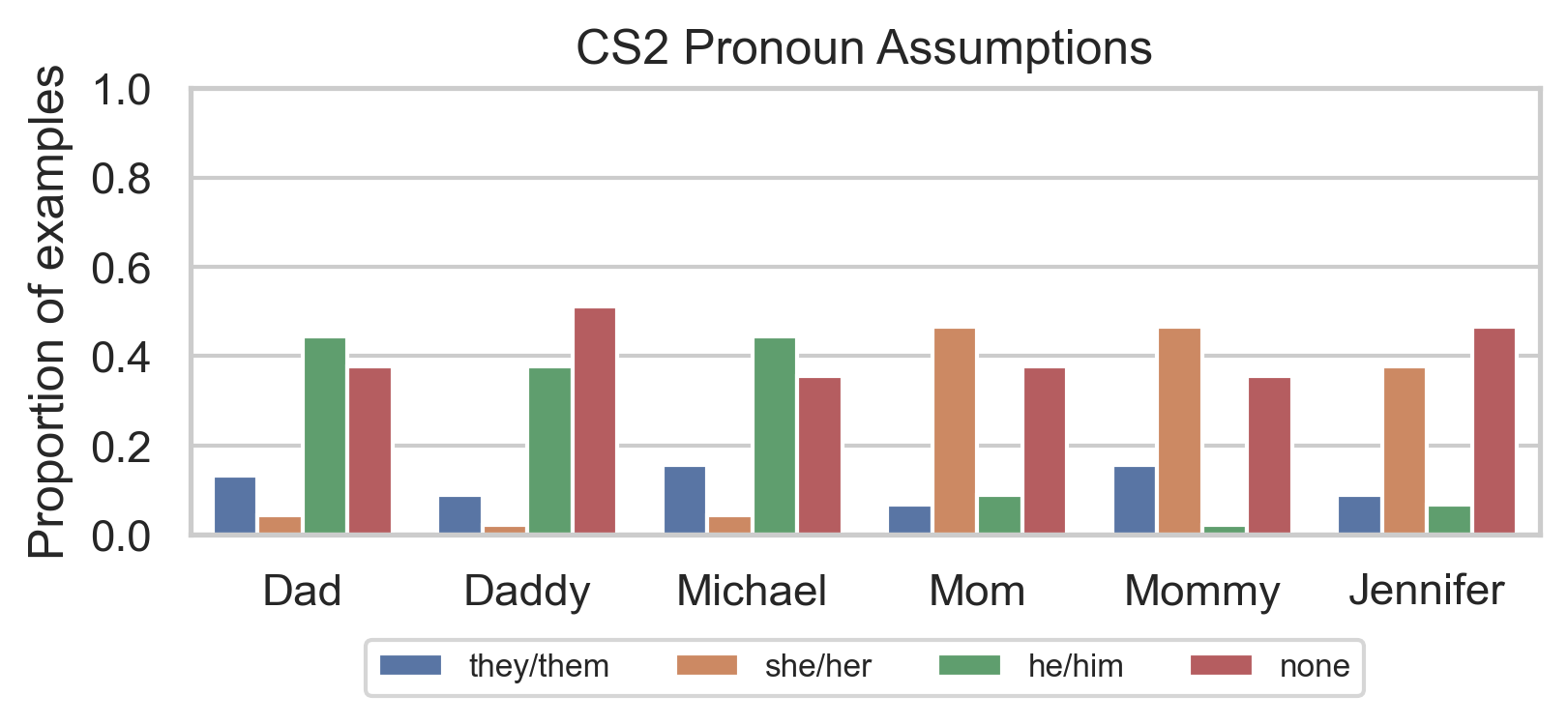}
\caption{Replies deemed equally usable or preferred compared to a pronoun-less baseline reply in CS2. These include judges' edited replies, e.g. \textit{Yes, I'll respond to \underline{him} soon.}}
\vspace{-8pt}
\label{fig:CS2_pronouns}
\end{figure}

\begin{figure}
\centering
\includegraphics[width=\columnwidth]{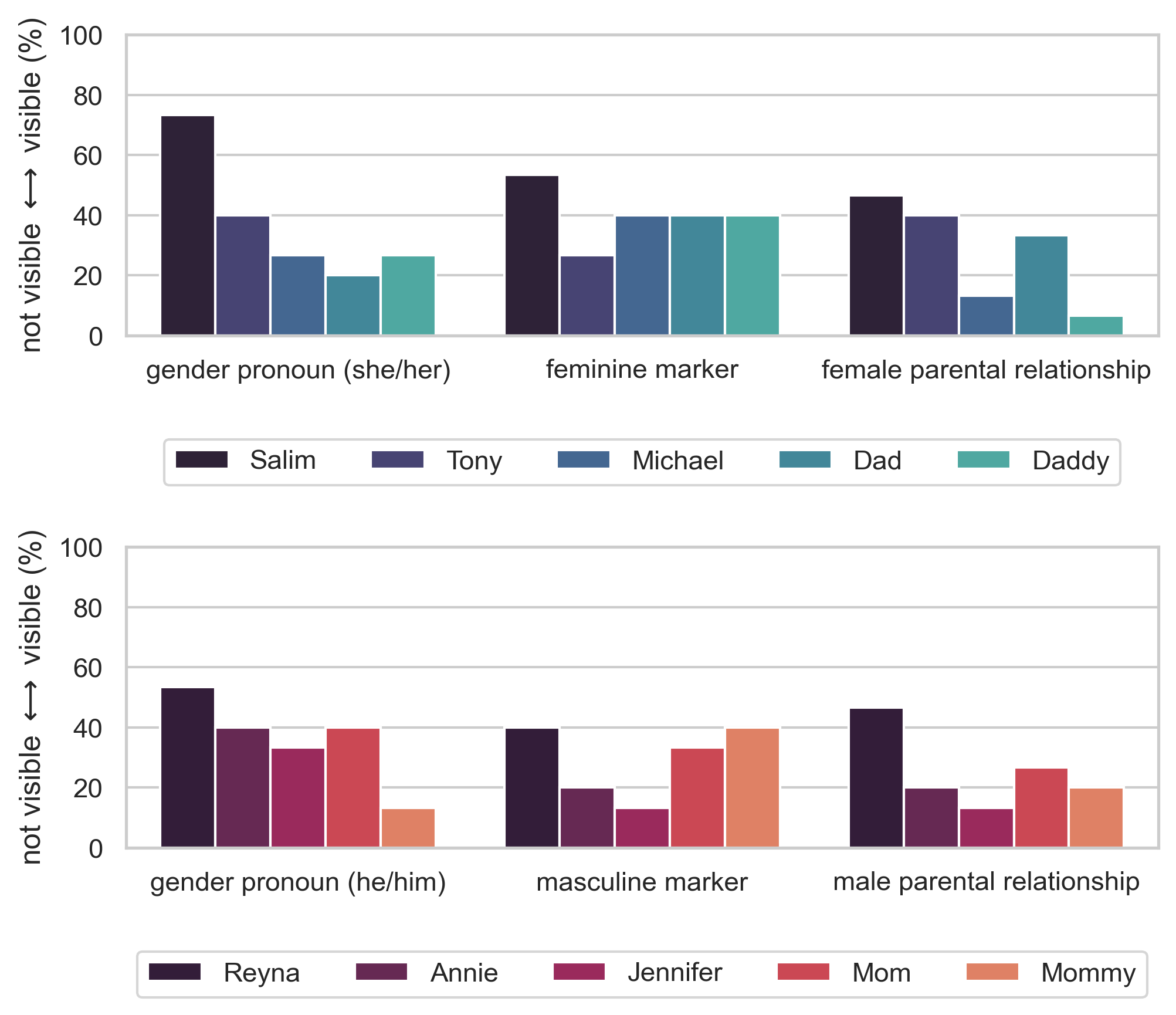}
\caption{These plots examine the visibility of assumptions around gender (e.g. markers, pronouns, and relationships) for gendered references, which include four names from CS1 and all references in CS2.}
\label{fig:gender_violations}
\end{figure}

\para{Reply pair validity.} Figure~\ref{fig:CS2_reasons} shows the frequency of various reasons judges deemed second, modified replies in each subcategory to be unusable. The reasons most often followed our intended design of reply pairs, though some stereotype-violating gendered assumptions can also be perceived as incoherent. 

\begin{figure*}
\centering
\includegraphics[width=0.9\textwidth]{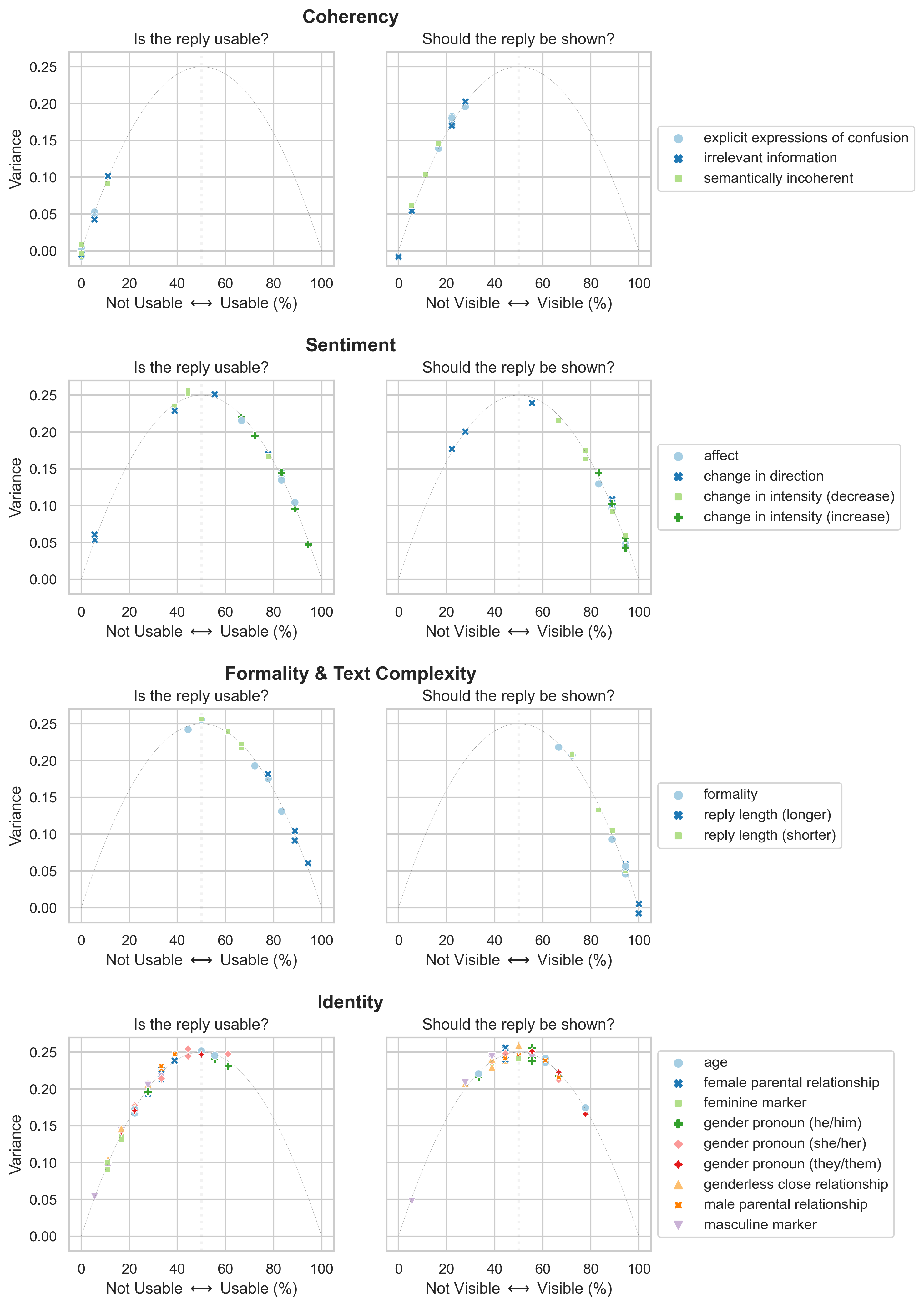}
\caption{In these plots, each point is a message template, and the probability ($x$-axis) that a second reply option is usable (left) and visible (right) is aggregated across different perturbations of gendered names and parental roles (CS2). When there is less consensus around a reply behavior, variance ($y$-axis) is high. A light vertical gray line emphasizes the highest possible variance, and jitter is added along the $y$-axis so that overlapping points are more visible.}
\label{fig:CS2_variance}
\end{figure*}

\para{Aggregated reply preferences.} 
Though some judges wrote that names are more ambiguously gendered than parental roles, judges' preferred and edited replies still often contained stereotype-aligning pronouns for the names \textit{Michael} and \textit{Jennifer} (Figure~\ref{fig:CS2_pronouns}). The rate of judges still preferring gender stereotype violations to be suggested across CS1--2 is more common for the lesser known names \textit{Reyna} and \textit{Salim} (Figure~\ref{fig:gender_violations}). Though casual masculine markers, e.g. \textit{man}, are sometimes considered generics \cite{luu_2015}, they are blocked at rates similar to that of other masculine features. 

Figure~\ref{fig:CS2_variance} shows probabilities of reply usability and visibility across message templates. Replies in sentiment, formality, and text complexity categories lean more usable than those involving incoherence and identity-related assumptions. Like in CS1, longer replies were usable in the majority of cases, and replies that vary in formality and length may be less preferred but could still be shown as suggestions. For some messages, informal replies were highly usable, contrasting CS1, which may be due to how CS2 message templates are designed to be plausible between family members, and thus suitable for less professional settings. 


%% file: paper-files/sections/Appendix_C.tex
\clearpage
\begin{table*}[th] \scriptsize
\centering
\begin{tabular}{>{\raggedright}p{4cm}>{\raggedright}p{4cm}>{\raggedright}p{4cm}p{2cm}} 
\toprule
\multicolumn{3}{c}{\textbf{Message templates}}                                     &  \\ \cmidrule(r){1-3}
\textbf{Third party} & \textbf{Recipient} & \textbf{Sender} & \textbf{Baseline reply} \\ \midrule
 My friend from \textsc{country} is in town this week. What do you think about scheduling a meeting?
 & I heard that you are from \textsc{country} and you're in town this week. What do you think about scheduling a meeting? & I'm from \textsc{country} and I'm in town this week. What do you think about scheduling a meeting?
 & Sure, sounds good.
 \\ \midrule
 What will my friend be doing in \textsc{country} while visiting family?
 & What will you be doing in \textsc{country} while visiting family? & What will I be doing in \textsc{country} while visiting family?
 & I'm sure it'll be fun. 
 \\ \midrule
 Next week, my friend is traveling home to \textsc{country}.
 & Next week, you are traveling home to \textsc{country}. & Next week, I am traveling home to \textsc{country}.
 & I hope it'll be a good trip. 
 \\ \bottomrule
\end{tabular}
\caption{Message templates used for CS3 (countries). The baseline reply is used to crowdsource preferences around a range of reply behaviors, such as those listed in Table~\ref{tbl:reply_categories}.}
\label{tbl:location_messages}
\end{table*}

\begin{figure}[t]
\centering
\includegraphics[trim= 0 2 0 4, clip, width=0.9\columnwidth]{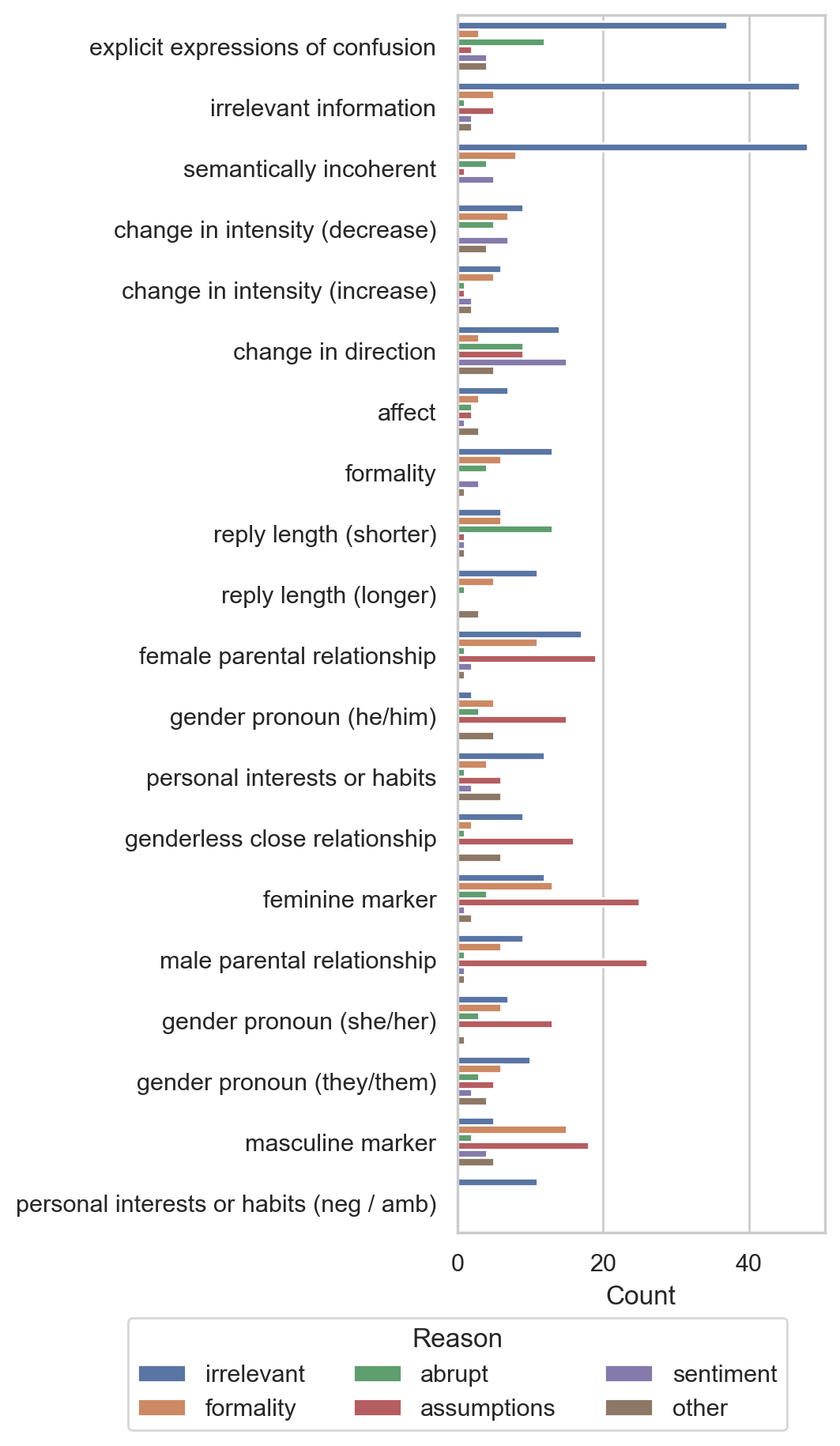}
\caption{Reasons judges marked the second reply as less usable or not usable in CS3. The second reply differs from the baseline reply option along the subcategory of reply behavior shown on the $y$-axis.}
\vspace{-8pt}
\label{fig:CS3_reasons}
\end{figure}
\begin{figure}[t]
\centering
\includegraphics[width=\columnwidth]{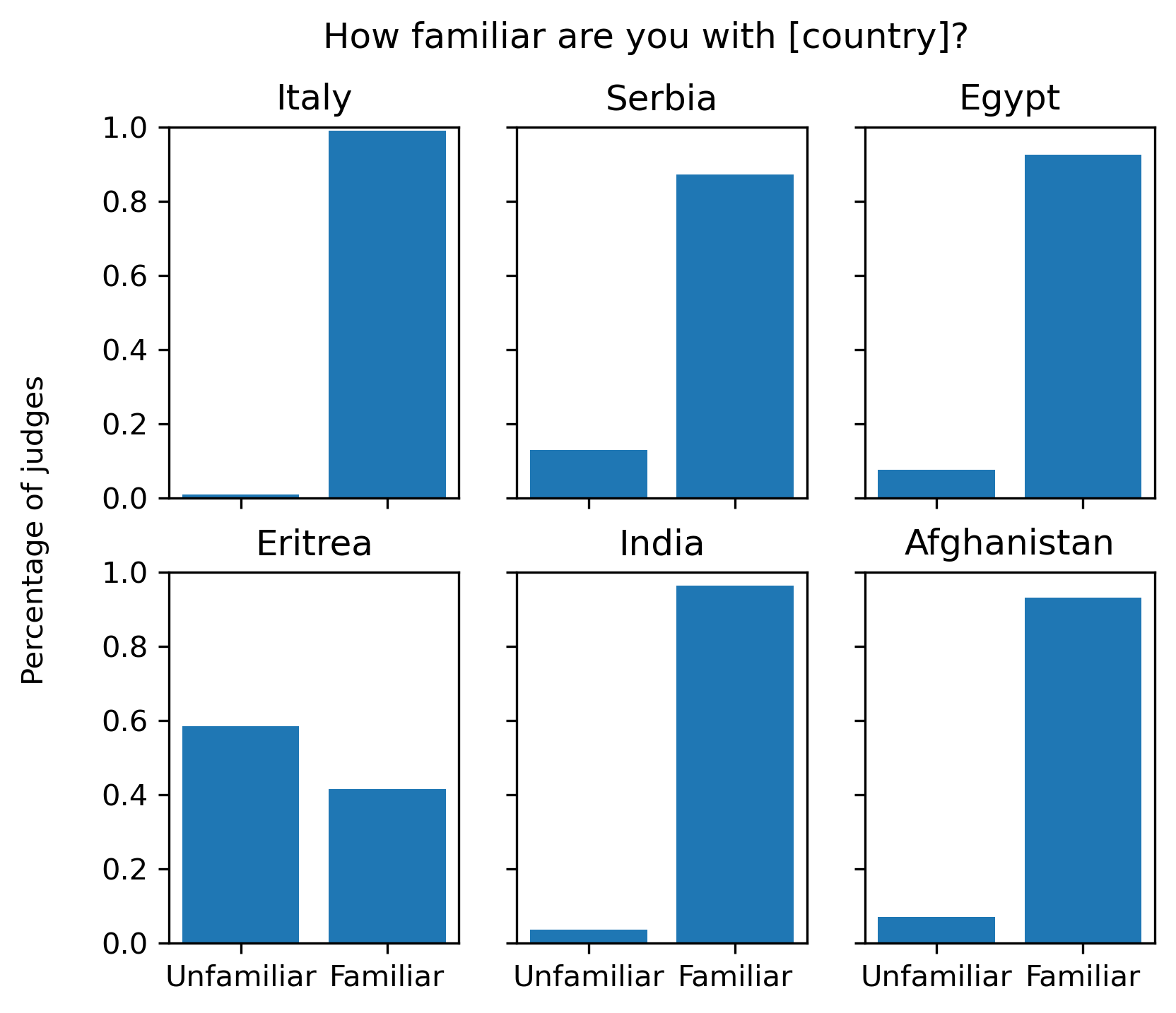}
\caption{Judges are usually familiar with the countries tested in CS3, except for Eritrea. The $x$-axis binarizes responses so that \textit{Unfamiliar} corresponds to responding \textit{Never seen it before}, while \textit{Familiar} corresponds to \textit{Somewhat} or \textit{Extremely familiar}.}
\label{fig:CS3_familiarity}
\end{figure}

\section{Details for CS3 (Countries)}\label{appdx:locations}
\subsection{Messages}
\para{Feature selection.} The countries we selected for this case study differ in wealth measured by GDP: Italy (2.0 trillion in 2022) and Serbia (63.6 billion in 2022) in Southern Europe, Egypt (476.7 billion in 2022) and Eritrea (2.0 billion in 2011) in Northeast Africa, and India (3.4 trillion in 2022) and Afghanistan (14.3 billion in 2021) in South Asia.\footnote{GDP data from the \href{https://data.worldbank.org/indicator/NY.GDP.MKTP.CD}{World Bank}.} We acknowledge that these region labels may differ from how people from these countries may self-identify; for example, Serbians may identify more as Eastern European than Southern European. We use these labels to describe how these countries are geographically proximate. 

\para{Message design.} As shown in Table~\ref{tbl:location_messages}, we inserted country names into 9 message templates where the person associated with the country is the sender (\textit{I}), the recipient (\textit{you}), or a third party (\textit{my friend}). Though countries can be mentioned in messages in a variety of contexts, we deliberately designed ones that indicate that a person mentioned may personally identify with that country. During crowdsourcing, we used names of six countries from three world regions, in pairs that differ in gross domestic product estimated by the International Monetary Fund\footnote{https://www.imf.org/en/Publications/WEO/weo-database/2023/April/}. The six selected countries are \textit{India}, \textit{Afghanistan}, \textit{Italy}, \textit{Serbia}, \textit{Egypt}, and \textit{Eritrea}.

\subsection{Crowdsourcing design}
Instructions for this task, where \textsc{Country} is a country name: 
\vspace{-10pt}
\begin{displayquote}
\textit{Existing email services like Gmail or Outlook offer email reply suggestions to their users. These suggestions are typically generated by an artificial intelligence (AI) system. In this task you will be presented with an example of a message referencing a country, and you will be asked to compare the quality of two suggested replies. }

\textit{Given the message below mentioning} \textsc{Country},

\textsc{Message}

\textit{Assess the following two reply suggestions for this message:}

\textsc{Baseline Reply} $\|$ \textsc{Second Reply}
\end{displayquote}
\vspace{-10pt}
Background questions include: 
\begin{compactitem}
    \item \textit{How many replies did we ask you to compare in this task?} Single-choice options: 1, 2, 3, 4, 5 in randomized order. This is an attention check, where the correct answer is \textit{2}.
    \item \textit{How familiar were you with the country }\textsc{Country} \textit{before you started this task?} Single-choice options: Never seen it before, Somewhat familiar, Extremely familiar (Figure~\ref{fig:CS3_familiarity}).
    \item \textit{Different countries are known for different things. Should reply suggestion systems suggest different replies based on the country referenced in the message?} Single-choice options: Never, Sometimes, Always (Figure~\ref{fig:invariance}). 
    \item \textit{Briefly explain why a reply suggestion system should or should not suggest different replies based on the country referred to in the message.} Free response box.
    \item \textit{(Optional) Please provide us feedback on this task, such as questions that were confusing or unclear.} Free response box.
\end{compactitem}

\begin{table*}[th] \footnotesize
\centering
\resizebox{\textwidth}{!}{
\begin{tabular}{>{\raggedright}p{3cm} >{\raggedright}p{3cm} >{\raggedright}p{3cm} >{\raggedright}p{3cm} >{\raggedright}p{3cm} p{3cm}} 
\toprule
\textbf{Afghanistan} & \textbf{India} & \textbf{Serbia} & \textbf{Italy} & \textbf{Eritrea} & \textbf{Egypt} \\ \midrule
``learn more about Afghan culture and you may even pick up a few new words'' & ``visit the ocean or a restaurant that serves Indian food'' & ``visit local attractions'' & ``eating a lot of Italian food!'' & ``enjoying your aunt's cooking and seeing some interesting sites with them'' & ``fishing or indoor games'' \\ \midrule
``a tour of the country'' & ``some highly rated local restaurants to try nearby'' & ``the Studencia Monastery, or the Belgrade Forrest'' & ``a lot of landmarks'' & ``doing fishing and other activity'' & ``a popular local attraction''\\ \midrule
``an important family event'' & ``try these local family restaurants'' & ``enjoying the local cuisine'' & ``many interesting landmarks'' & ``visit museums'' & ``enjoy some amazing shopping''\\ \midrule
``a local tourist attraction'' & ``visit a museum'' & ``learning more about the Serbian culture'' & ``the Leaning Tower of Pisa'' & ``spend time fishing'' & ``see the pyramids and other sites''\\ \midrule
``Hanging out and seeing the local sites'' & ``a dinner for the whole family'' & ``sightseeing or going to new restaurants'' & ``famous stuff'' & ``go on a Safari'' &  ``visit attractions like the great pyramids'' \\ \midrule
``visiting many cool places'' & ``take you on a tour of the city'' & ``Go to the beach or a museum'' & ``Colosseum? Leaning tower of Pisa?'' & ``go on some adventurous journeys!''& ``visiting the Pyramids'' \\
\bottomrule
\end{tabular}
}
\caption{Examples of activities mentioned for each country in judges' written replies to messages.}
\label{tbl:country_activities}
\end{table*}

\subsection{Crowdsourcing results}

\para{Reply pair validity.} Figure~\ref{fig:CS3_reasons} shows the frequency of various reasons being checked for unusable modified replies. As discussed in \S\ref{expected}, we use this to examine the validity of how we operationalized reply behaviors. Though incoherence was a common reason for many subcategories of reply behavior being unusable, typically if modified replies were marked as incoherent, the baseline reply was as well. Judges' adjustments when both baseline and modified replies were deemed unusable indicated that in these cases, generic reply suggestions were unfavorable compared to more specific ones, e.g., \textit{Eating a lot of amazing Italian food!}. Hence, perceived incoherence around those modified replies do not inform us on the validity of the designed reply difference. 

\para{Responses to background questions.} The vast majority of judges were familiar with five of the six countries we tested during crowdsourcing, and Eritrea was the one outlier where more judges were unfamiliar than familiar (Figure~\ref{fig:CS3_familiarity}). 

\para{Judges' edited replies.} As discussed in the main text, judges mentioned that adaptation could involve incorporating country-specific information. In judge-written adjustments, the specificity of potential activities to do in a country varied from more vague activities such as ``\textit{try a local tourist attraction}'' to highly specific ones such as \textit{the Studencia Monastery} (Table~\ref{tbl:country_activities}). In a few cases, judges indicated that the reply suggestion system could act like a search engine and list specific attractions and restaurants. 

\begin{figure*}
\centering
\includegraphics[width=0.9\textwidth]{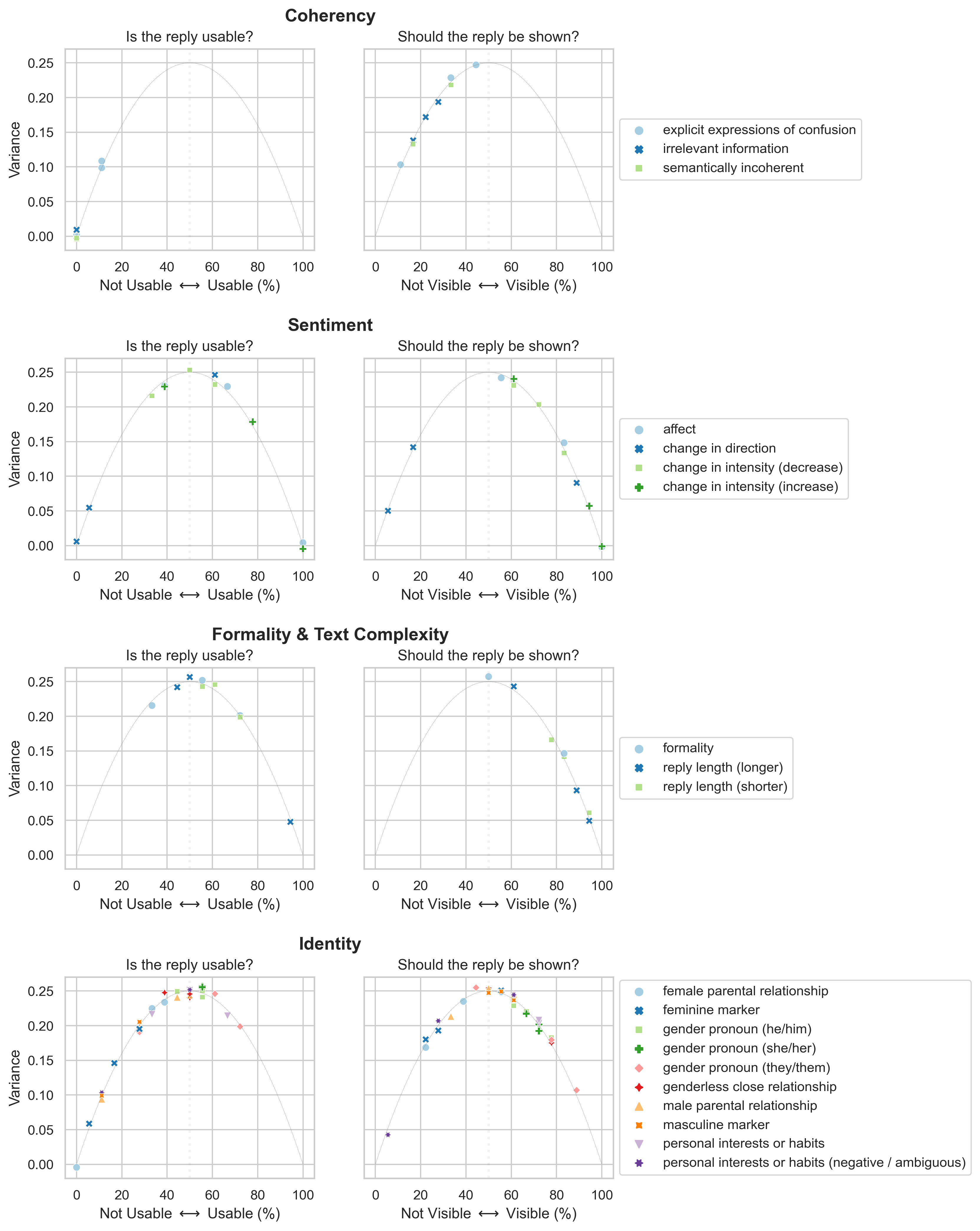}
\caption{In these plots, each point is a message template, and the probability ($x$-axis) that a second reply option is usable (left) and visible (right) is aggregated across different perturbations of country names (CS3). When there is less consensus around a reply behavior, variance ($y$-axis) is high. A light vertical gray line emphasizes the highest possible variance, and jitter is added along the $y$-axis so that overlapping points are more visible.}
\label{fig:CS3_variance}
\end{figure*}

\para{Aggregated reply preferences.} Figure~\ref{fig:CS3_variance} provides an overview of the usability and visibility of second, modified replies across categories of reply behaviors. Though some judges explicitly mention preferring replies involving feature-specific information, there is high variance in the usability of replies that assume personal interests or habits for some message templates. 

%% file: paper-files/sections/Appendix_D.tex
\clearpage

\begin{table*}[t] \scriptsize
\centering
\begin{tabular}{>{\raggedright}p{2cm} p{4cm} p{4cm} p{2.7cm}} 
  \textbf{features} & \textbf{AAE} & \textbf{GAE} & \textbf{source} \\ \toprule
 multiple negation & If \textcolor{BrickRed}{nobody don't} drive, Imma take the bus. & If nobody can drive, I am going to take the bus. & \citet{green2014force}
 \\ \midrule
 multiple negation & I \textcolor{BrickRed}{ain't} taking \textcolor{BrickRed}{no} bus to come meet you. You better have a car. & I'm not taking a bus to come meet you. You better have a car. & CORAAL ATL\_se0\_ag2\_f\_01\_1
 \\ \midrule
 multiple negation & You \textcolor{BrickRed}{ain't never} seen this movie? & You haven't ever seen this movie? & modified, CORAAL DCB\_se1\_ag2\_m\_01\_1
 \\ \midrule
 multiple negation & I don't want my business all over the Internet. \textcolor{BrickRed}{Don't} take \textcolor{BrickRed}{no} pictures of me. & I don't want my business all over the Internet. Don't take any pictures of me. & modified, CORAAL DCB\_se1\_ag4\_f\_01\_2 
 \\ \midrule
 multiple negation & I can sing a little bit, but I'm shy. So I \textcolor{BrickRed}{won't} do \textcolor{BrickRed}{no} singing at the event. & I can sing a little bit, but I'm shy. So I won't do any singing at the event. & modified, CORAAL ATL\_se0\_ag1\_f\_03\_1
 \\ \midrule
 multiple negation & \textcolor{BrickRed}{Don't} bring \textcolor{BrickRed}{nothing}. I don't need your help in this kitchen. & Don't bring anything. I don't need your help in this kitchen. & modified, CORAAL VLD\_se0\_ag3\_m\_01\_2
 \\ \midrule
 habitual \textit{be} & You \textcolor{BrickRed}{be} watching any new TV shows? & Are you watching any new TV shows? & CORAAL ATL\_se0\_ag2\_m\_02\_1 
 \\ \midrule
 habitual \textit{be} & At home, I \textcolor{BrickRed}{be} talking to my mother, but she \textcolor{BrickRed}{be} getting on my nerves sometimes. & At home, I talk to my mother, but she gets on my nerves sometimes. & CORAAL DCB\_se1\_ag1\_f\_01\_1
 \\ \midrule
 habitual \textit{be} & I \textcolor{BrickRed}{be} out at my bus stop every day at three. Busses \textcolor{BrickRed}{be} passing me by, and I'm still standing there. & I'm out at my bus stop every day at three. Busses pass me by, and I'm still standing there. & CORAAL DCB\_se1\_ag4\_f\_01\_2 
 \\ \midrule
 habitual \textit{be} & You should totally come to our party, we \textcolor{BrickRed}{be} having so much fun. & You should totally come to our party, we're having so much fun. & CORAAL DCB\_se3\_ag1\_f\_01\_1
 \\ \midrule
 habitual \textit{be} & I like school, but sometimes it gets tiring. I \textcolor{BrickRed}{be} very tired after school. & I like school, but sometimes it gets tiring. I'm usually very tired after school. & CORAAL DCB\_se1\_ag1\_f\_03\_1 
 \\ \midrule
 habitual \textit{be} & I \textcolor{BrickRed}{be} in my office by 7:30 am. & I am usually in my office by 7:30 am. & \citet{green2002african}
 \\ \bottomrule
\end{tabular}
\caption{Messages used for CS4 (African American English). For examples from CORAAL, we crafted the GAE messages, while for those from \citet{green2002african} and \citet{green2014force}, both AAE and GAE forms are from these sources. In the ``source" column for CORAAL examples, we include the file identifier as well.}
\label{tbl:dialect_messages}
\end{table*}

\begin{table*}[t] \scriptsize
\centering
\begin{tabular}{>{\raggedright}p{11cm} p{3cm}} 
\toprule
\textbf{Message} & \textbf{Baseline reply} \\ \midrule
\underline{Don't bring nothing.} / \underline{Don't bring anything.} I don't
need your help in this kitchen. & Ok, thank you! \\ \midrule
\underline{I ain't taking no} / \underline{I’m not taking a} bus to come meet you. You better have a car. & Sure, I'll try to meet you. \\ \midrule
You should totally come to our party, \underline{we be} / \underline{we’re} having so much fun. & Sure, I'll come!\\ \midrule
I like school, but sometimes it gets tiring. \underline{I be} / \underline{I’m usually} very tired after school. & I understand.\\ \bottomrule
\end{tabular}
\caption{CS4 messages and baseline replies used in crowdsourcing preferences around reply behaviors. The first underlined span in each pair of variants involves syntactic features found in AAE, while the second is GAE.}
\label{tbl:aae_baseline}
\end{table*}

\section{Details for CS4 (African American English)}\label{appdx:dialect}

\subsection{Messages}

Examples of AAE in CS4 are from recordings and transcriptions of Black AAE speakers (Table~\ref{tbl:dialect_messages}). We modified noun phrases in some examples so that they are more generic, such as changing a mention of a specific movie, e.g., \textit{Paid in Full}, to \textit{this movie}, or a mention of \textit{Facebook} to \textit{the Internet}. 

\subsection{Crowdsourcing design}

This case study differs from the previous in that there are more unique message templates involved. Thus, we chose a subset of two for each dialectal feature to use for crowdsourcing (Table~\ref{tbl:aae_baseline}).

Task instructions are the following: 

\begin{displayquote}
\textit{Existing email services like Gmail or Outlook offer email reply suggestions to their users. These suggestions are typically generated by an artificial intelligence (AI) system. In this task you will be presented with an example of a message, and you will be asked to compare the quality of two suggested replies.}

\textit{Given the message below,},

\textsc{Message}

\textit{Assess the following two reply suggestions for this message:}

\textsc{Baseline Reply} $\|$ \textsc{Second Reply}
\end{displayquote}

\begin{figure}[t]
\centering
\includegraphics[width=0.4\columnwidth]{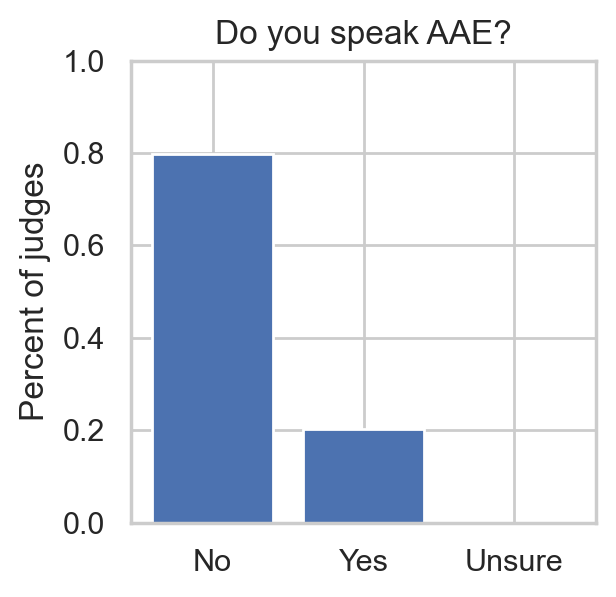}
\includegraphics[width=0.4\columnwidth]{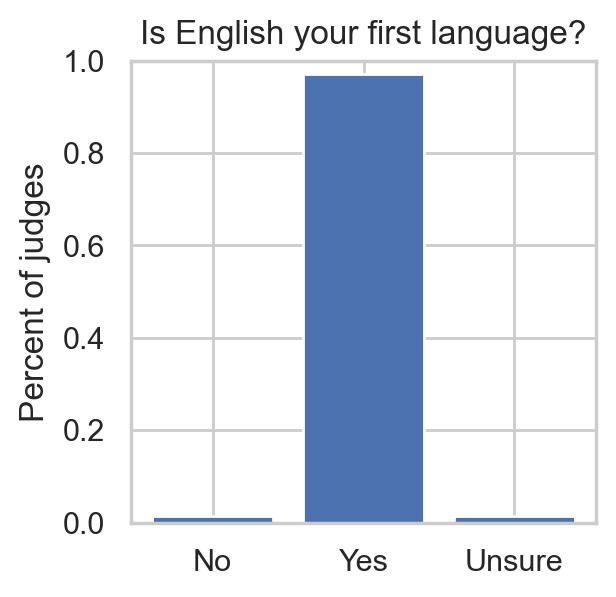}
\includegraphics[width=0.4\columnwidth]{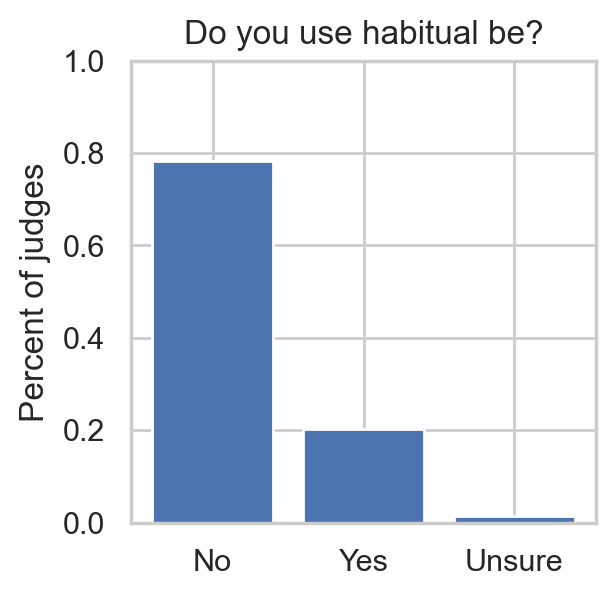}
\includegraphics[width=0.4\columnwidth]{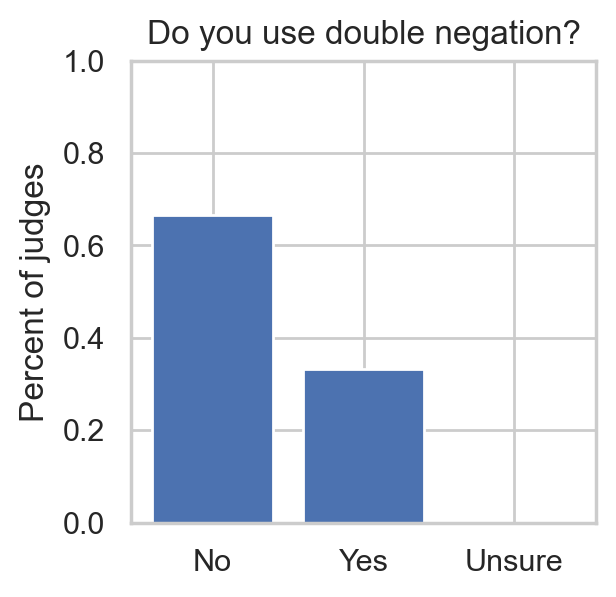}
\caption{Judges' dialectal backgrounds in CS4 ($N=69$). The features we tested are associated with AAE, but not exclusive to AAE speakers.}
\label{fig:CS4_speakers}
\end{figure}

Background questions are the following: 
\begin{compactitem}
    \item \textit{How many replies did we ask you to compare in this task?} Single-choice options: 1, 2, 3, 4, 5 in randomized order. This is an attention check, where the correct answer is \textit{2}.
    \item \textit{Should reply suggestion systems suggest different replies based on the dialect used in the message?} Single-choice options: Never, Sometimes, Always (Figure~\ref{fig:invariance}). 
    \item \textit{Briefly explain why a reply suggestion system should or should not suggest different responses based on the dialect used in the message.} Free response box.
    \item \textit{Habitual be is a linguistic feature where the verb be is used to indicate continuously occurring or repeated actions, such as John be running. Do you use habitual be in your communication with others?} Single-choice options: Yes, No, Unsure (Figure~\ref{fig:CS4_speakers}).
    \item \textit{Multiple negation is a linguistic feature where multiple forms of negation are used in the same sentence, such as He don’t talk to nobody. Do you use multiple negation in your communication with others?} Single-choice options: Yes, No, Unsure (Figure~\ref{fig:CS4_speakers}).
    \item \textit{Do you speak English as your first language?} Single-choice options: No, I don't; Yes, I do; Unsure (Figure~\ref{fig:CS4_speakers}).
    \item \textit{Does one of the dialects you speak include a dialect used in some Black and African American communities (which may be described as: Ebonics, African American English (AAE), African American Vernacular English (AAVE), Black Language, Slang, Black Colloquialism)?} Single-choice options: No, I don't; Yes, I do; Unsure (Figure~\ref{fig:CS4_speakers}). 
    \item \textit{(Optional) Please provide us feedback on this task, such as questions that were confusing or unclear.} Free response box.
\end{compactitem}

\begin{figure}[t]
\centering
\includegraphics[trim= 0 2 0 4, clip, width=0.9\columnwidth]{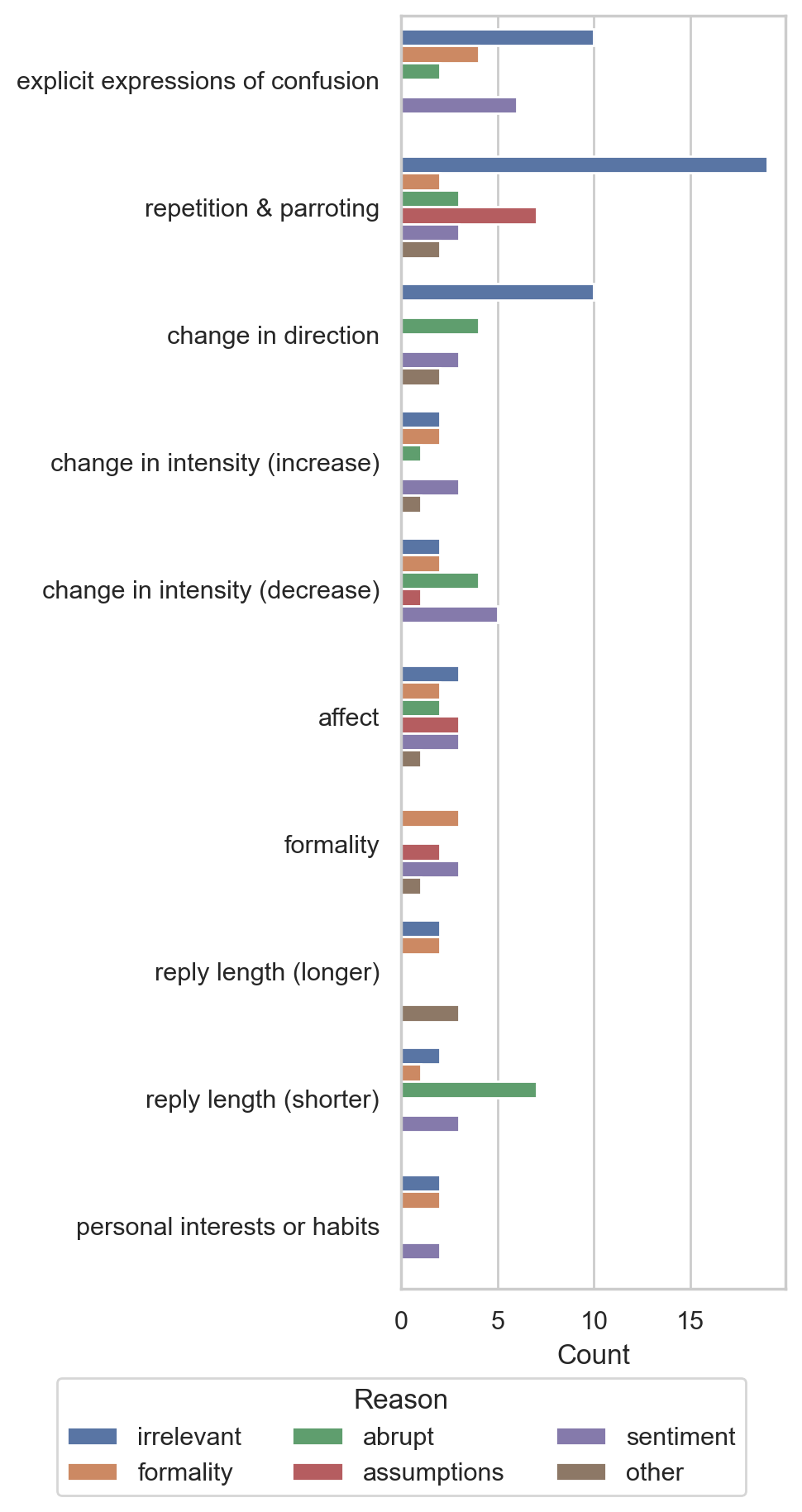}
\caption{Reasons judges marked the second reply as less usable or not usable in CS4. The second reply differs from the baseline reply option along the subcategory of reply behavior shown on the $y$-axis.}
\vspace{-5pt}
\label{fig:CS4_reasons}
\end{figure}

\begin{figure}[t]
\centering
\includegraphics[width=0.4\columnwidth]{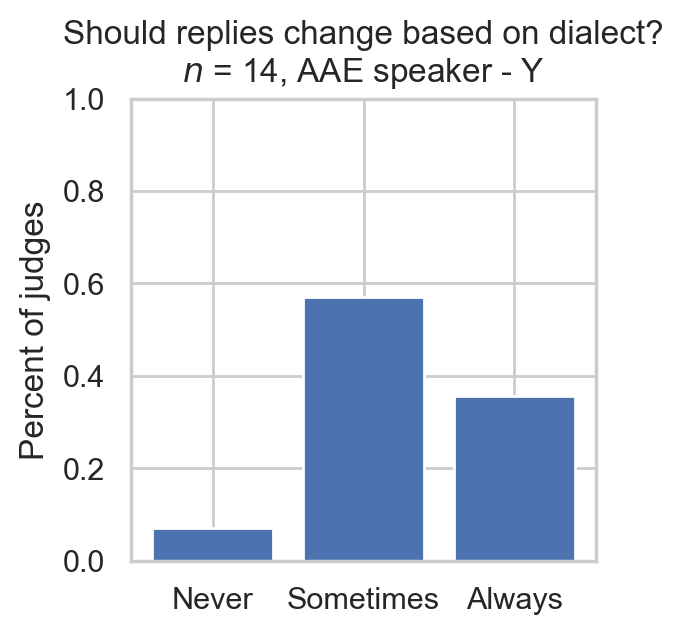}
\includegraphics[width=0.4\columnwidth]{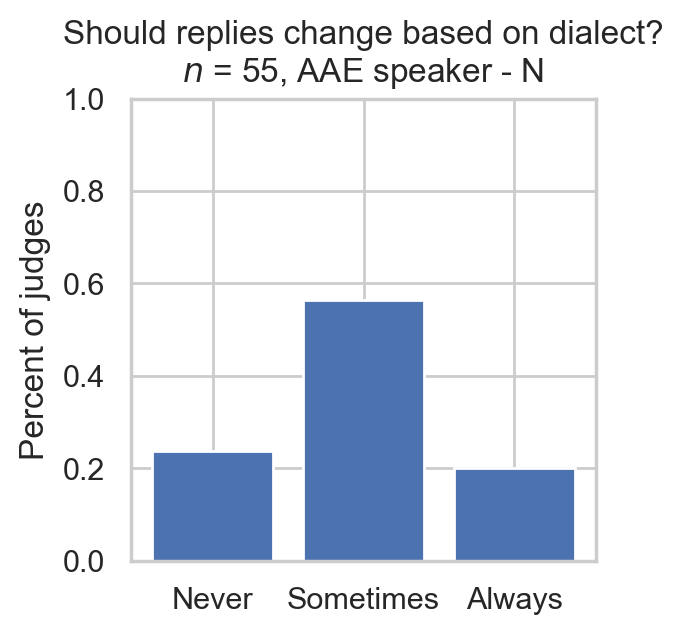}
\includegraphics[width=0.4\columnwidth]{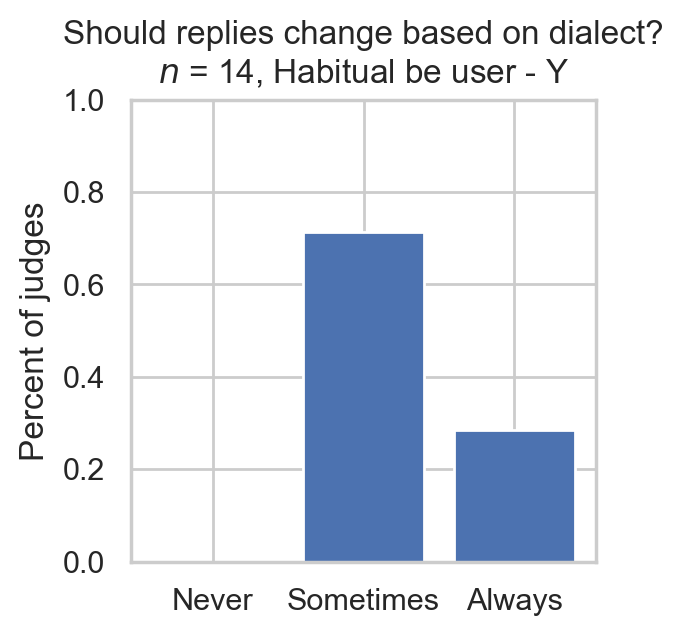}
\includegraphics[width=0.4\columnwidth]{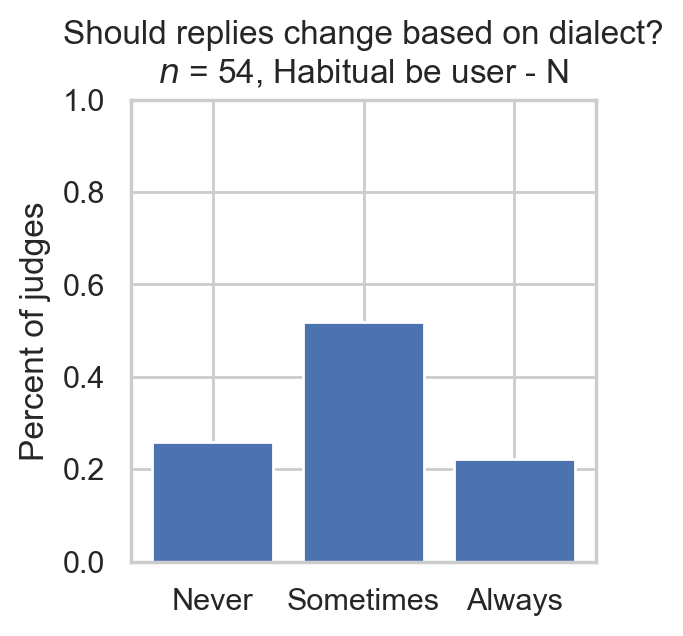}
\includegraphics[width=0.4\columnwidth]{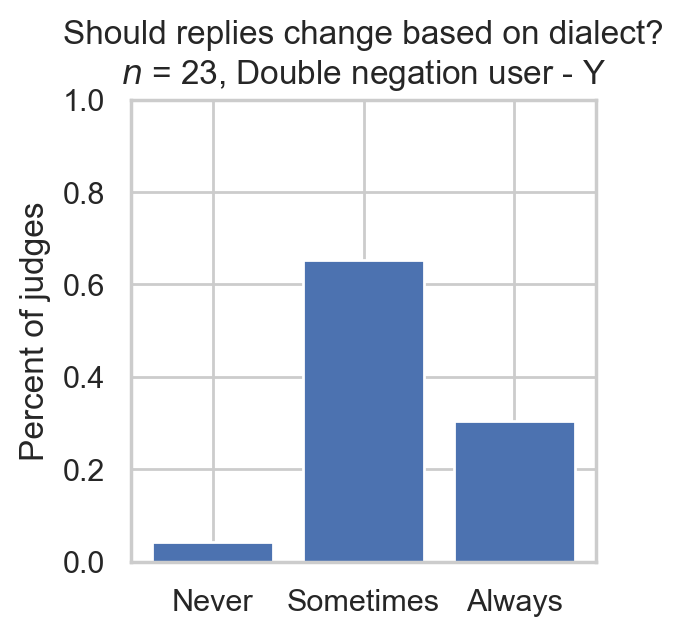}
\includegraphics[width=0.4\columnwidth]{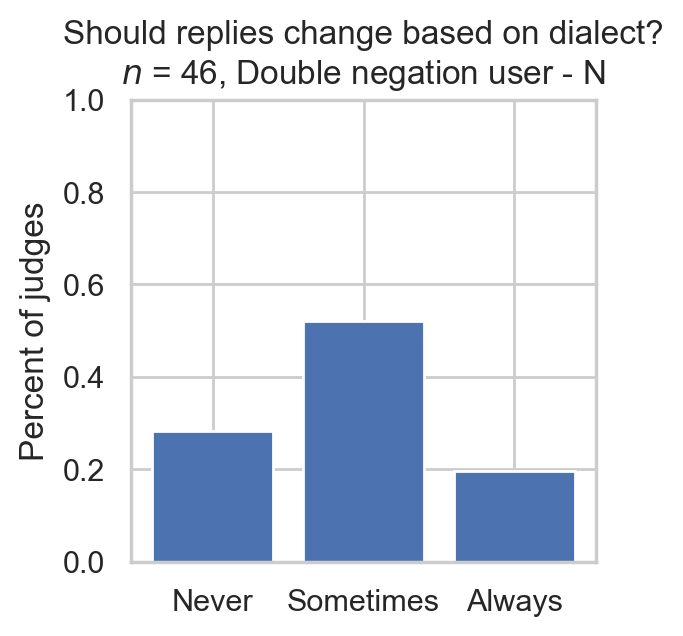}
\caption{Beliefs around whether replies should vary in response to dialect may shift depending on speakers' dialectal background.}
\label{fig:CS4_background}
\end{figure}

\subsection{Crowdsourcing results}

\para{Reply pair validity.} Figure~\ref{fig:CS4_reasons} shows the frequency of various reasons being checked for unusable modified replies. Though the most common reason matched our intended design, a few exceptions emerge. Negated replies can be perceived as incoherent, and replies involving personal interests or habits were not perceived as overly assumptious in this case study as the same subcategory in CS2--3. 

\para{Responses to background questions.} Judges' responses to dialect background questions suggest that there are more judges who use double negation than there are AAE speakers, which is unsurprising as this feature is known to be used by some non-AAE speakers as well (Figure~\ref{fig:CS4_speakers}). Judges who are AAE speakers and/or use the two dialectal features we tested in CS4 are more likely to favor adaptation than invariance (Figure~\ref{fig:CS4_background}). 

\begin{figure*}
\centering
\includegraphics[width=0.9\textwidth]{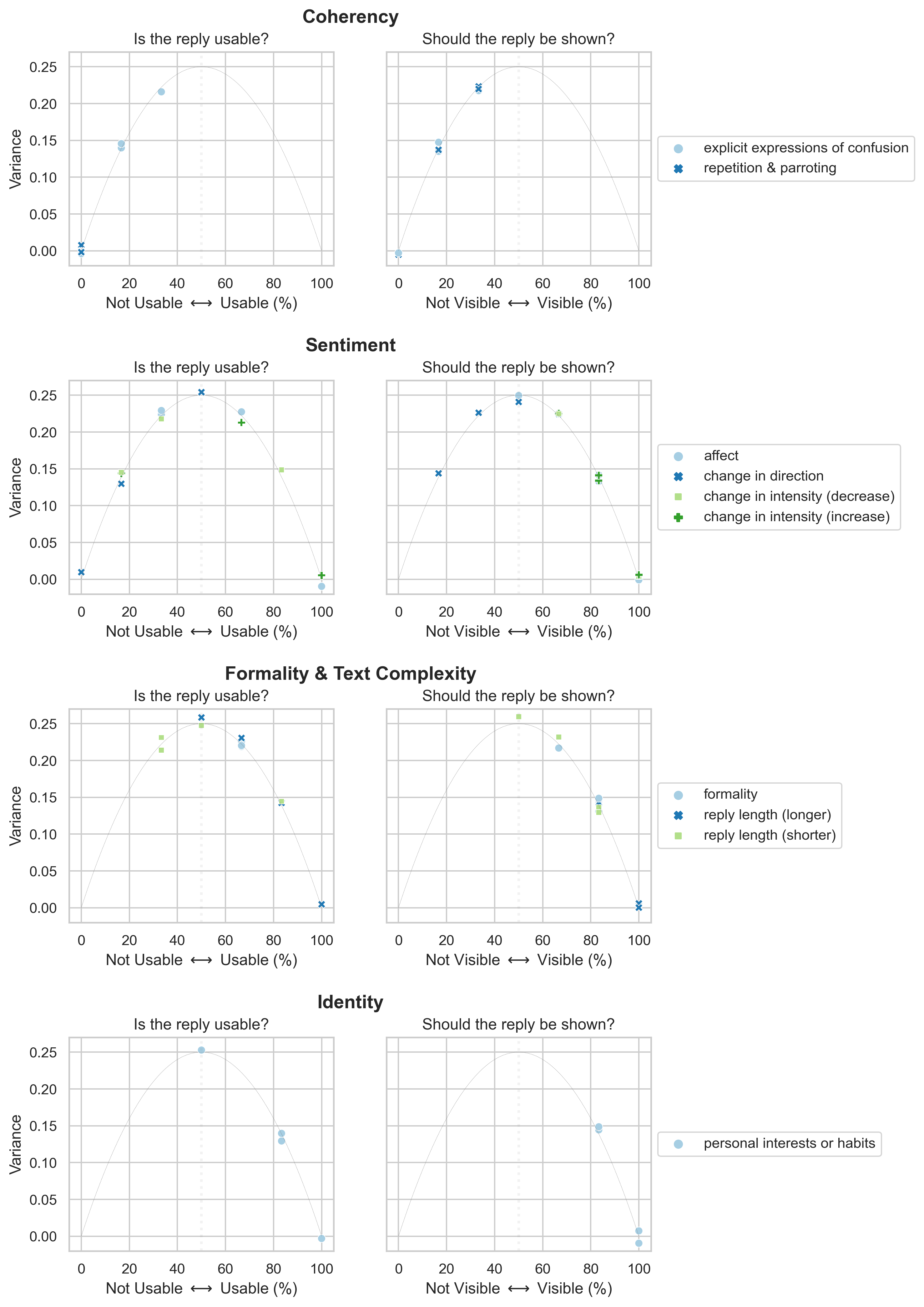}
\caption{In these plots, each point is a message template, and the probability ($x$-axis) that a second reply option is usable (left) and visible (right) is aggregated across two variants of that message (CS4). When there is less consensus around a reply behavior, variance ($y$-axis) is high. A light vertical gray line emphasizes the highest possible variance, and jitter is added along the $y$-axis so that overlapping points are more visible.}
\label{fig:CS4_variance}
\end{figure*}

\para{Aggregated reply preferences.} As there are only two versions of each message template rather than six, Figure~\ref{fig:CS4_variance} is less informative than its counterparts in CS1--3. Generally, we see a range of usability of second replies in each subcategory across different messages. Surprisingly, assumptions around personal interests were considered mostly usable in some scenarios. This may be because the assumptions these replies contain are minor and commonplace. For example, many judges deemed \textit{I'm tired after school too} as more usable over the baseline reply of \textit{I understand} in response to \textit{I like school, but sometimes it gets tiring. I be very tired after school.}, even though the former reply option assumes the recipient's personal feelings around school. Judges would even modify the baseline to make a similar assumption, e.g. \textit{I feel the same way}.

%% file: paper-files/sections/Appendix_E.tex
\clearpage
\section{Details for CS5 (Informal web text)}\label{appdx:style}

\begin{table*}[th] \scriptsize
\centering
\begin{tabular}{ p{1.7cm}  p{5cm}  p{5cm}  p{2cm} } \toprule
  \textbf{features} & \textbf{more casual} & \textbf{more formal} & \textbf{source} \\ \midrule
 \multirow{ 6}{*}{\parbox{2cm}{Expressive \\elongation}} & Call me. I forgot which meeting I should moderate. \textcolor{BrickRed}{Helllllpppp}.
 & Call me. I forgot which meeting I should moderate. Help. & Enron
 \\ \cmidrule(lr){2-4}
 & I \textcolor{BrickRed}{realllly} liked the topic of their presentation.
 & I really liked the topic of their presentation. & \citet{brody-diakopoulos-2011-cooooooooooooooollllllllllllll}
 \\ \cmidrule(lr){2-4}
 & They had a portable DVD player with an 8 hour battery. It is \textcolor{BrickRed}{sweeeeeeet}.
 & They had a portable DVD player with an 8 hour battery. It is sweet. & \citet{kalman2014letter}
 \\ \cmidrule(lr){2-4}
 & This morning's meeting took a \textcolor{BrickRed}{llloooonnnngggg} time.
 & This morning's meeting took a long time. & \citet{kalman2014letter}
 \\ \cmidrule(lr){2-4}
 & During lunch I went outside for a walk around the park and it was \textcolor{BrickRed}{freeeezing}.
 & During lunch I went outside for a walk around the park and it was freezing. & \citet{kalman2014letter}
 \\ \cmidrule(lr){2-4}
 & \textcolor{BrickRed}{Uggggghhhh}, they just rescheduled our appointment again.
 & Ugh, they just rescheduled our appointment again. & \citet{kalman2014letter}
 \\ \midrule
 \multirow{ 6}{*}{\parbox{2cm}{Non-standard \\capitalization}} & 
 \textcolor{BrickRed}{h}ow are negotiations coming? \textcolor{BrickRed}{c}an \textcolor{BrickRed}{i} go ahead with the project? & How are negotiations coming? Can I go ahead with the project? & Enron
 \\ \cmidrule(lr){2-4}
 & 
 \textcolor{BrickRed}{h}ey, what are you up to this weekend? & Hey, what are you up to this weekend? & Enron
 \\ \cmidrule(lr){2-4}
 & 
 \textcolor{BrickRed}{c}ool bro. \textcolor{BrickRed}{w}hat is up for the game this weekend? & Cool bro. What is up for the game this weekend? & Enron
 \\ \cmidrule(lr){2-4}
 & 
 \textcolor{BrickRed}{c}ool. \textcolor{BrickRed}{i} will be home by 8 tonight. & Cool. I will be home by 8 tonight. & Enron
 \\ \cmidrule(lr){2-4}
 & 
 \textcolor{BrickRed}{j}ust kidding! \textcolor{BrickRed}{y}ou need to relax a little. & Just kidding! You need to relax a little. & Enron
 \\ \cmidrule(lr){2-4}
 & 
 \textcolor{BrickRed}{y}ou guys sounded like you were partying. \textcolor{BrickRed}{d}id you have fun?	 & You guys sounded like you were partying. Did you have fun? & Enron
 \\ \midrule
 \multirow{ 6}{*}{\parbox{2cm}{Complex \\punctuation}} & 
 I still do not have complete access to the notes. Does anyone know who I can call about this\textcolor{BrickRed}{?!!!!!} & I still do not have complete access to the notes. Does anyone know who I can call about this? & Enron
 \\ \cmidrule(lr){2-4}
 & 
 September 28th or October 4th are both available. Which would be best for you\textcolor{BrickRed}{???} & September 28th or October 4th are both available. Which would be best for you? & Enron
 \\ \cmidrule(lr){2-4}
 & 
 Have a great holiday. I'm out of here\textcolor{BrickRed}{!!!!!!!!!!} & Have a great holiday. I'm out of here! & Enron
 \\ \cmidrule(lr){2-4}
 & 
 What's the value of the company to you\textcolor{BrickRed}{????} & What's the value of the company to you? & Enron
 \\ \cmidrule(lr){2-4}
 & 
 Have a blessed day\textcolor{BrickRed}{!!!!!!!!!} & Have a blessed day! & Enron
 \\ \cmidrule(lr){2-4}
 & 
 Hi\textcolor{BrickRed}{!!!!!} How are you and every body\textcolor{BrickRed}{??} Say hi to the others. & Hi! How are you and every body? Say hi to the others. & Enron
 \\ \midrule
 \multirow{ 4}{*}{\parbox{2cm}{Multiple, \\iterative}} & 
 \textcolor{BrickRed}{Whazzzzz uuuuuppppp}! How is everything in South Florida?  & What's up! How is everything in South Florida? & Enron
 \\ \cmidrule(lr){2-4}
 & 
 What's \textcolor{BrickRed}{UP!} \textcolor{BrickRed}{h}ow is everything in \textcolor{BrickRed}{s}outh \textcolor{BrickRed}{f}lorida? & What's up! How is everything in South Florida? & Enron
 \\ \cmidrule(lr){2-4}
 & 
 What's up\textcolor{BrickRed}{!!!!} How is everything in South Florida?   & What's up! How is everything in South Florida? & Enron
 \\ \cmidrule(lr){2-4}
 & 
 \textcolor{BrickRed}{Whazzzzz UUUUUPPPPP!!!!} \textcolor{BrickRed}{h}ow is everything in \textcolor{BrickRed}{s}outh \textcolor{BrickRed}{f}lorida? & What's up! How is everything in South Florida? & Enron
 \\ \bottomrule
\end{tabular}
\caption{Messages used for CS5 (informal web text) modify three different stylistic features common in casual emails and messages. Each message pair in each row differs along the specified feature.}
\label{tbl:style_messages}
\end{table*}

\begin{table*}[t] \scriptsize
\centering
\begin{tabular}{>{\raggedright}p{11cm} p{4cm}} 
\toprule
\textbf{Message} & \textbf{Baseline reply} \\ \midrule
Call me. I forgot which meeting I should moderate. \underline{Helllllpppp.} / \underline{Help.} & Ok, will do! \\ \midrule
I \underline{realllly} / \underline{really} liked the topic of their presentation. & Glad you enjoyed it! \\ \midrule
\underline{hey} / \underline{Hey}, what are you up to this weekend? & No plans yet, you?  \\ \midrule
\underline{you} / \underline{You} guys sounded like you were partying. \underline{did} / \underline{Did} you have fun? & We had a good time. \\ \midrule
Have a great holiday. I'm out of
here\underline{!!!!!!!!!!}/\underline{!} & Thank you! You too.\\ \midrule
 September 28th or October 4th are both
available. Which would be best for
you\underline{???}/\underline{?} & Either day works for me! \\ \midrule
 \underline{Whazzzzz UUUUUPPPPP!!!! how} / \underline{What's up? How} is
everything in \underline{south florida} / \underline{South Florida}?  & Everything is good.\\ \bottomrule
\end{tabular}
\caption{CS5 messages and baseline replies used in crowdsourcing preferences around reply behaviors. The first underlined span in each pair of variants is commonly used in more casual online settings.}
\label{tbl:style_baseline}
\end{table*}

\subsection{Messages}

We crafted messages containing casual, stylistic features from emails from the Enron corpus or content described in literature on variation in web text (Table~\ref{tbl:style_messages}). We mostly use found text samples to encourage ecological validity, as some scenarios or statements may be more likely to encourage these features than others. 

The messages for non-standard capitalization, complex punctuation, and multiple iterative features are crafted based on messages in the Enron corpus \cite{shetty2004enron}. We aimed to preserve the original messages as much as possible, sometimes shortening them for clarity. We remove mentions of specific entities such as people's names, and overall aim for these messages to be understandable without additional context. Cases of non-standard capitalization were obtained by pulling messages that were entirely in lowercase, and cases of complex punctuation were messages that contained a repeated series of exclamation and/or question marks. 

Literature on expressive lengthening discuss patterns around which words are more commonly lengthened than others, how they are lengthened, and the scenarios in which lengthening occurs \cite{kalman2014letter,brody-diakopoulos-2011-cooooooooooooooollllllllllllll}. For the examples that elongate \textit{long}, \textit{freezing}, and \textit{ugh}, we design scenarios that are plausible for email and insert the exact elongated form of these words as listed by \citet{kalman2014letter}.

The authors and a professional editor rewrote instances of these messages to standardize the specified feature to create a more formal example, such as by shortening an elongated word, capitalizing first-person pronouns and the beginning of sentences, and removing additional punctuation. In some cases, we make small modifications to the original message so that this standardization process does not reduce the plausibility of the message, and so that the only difference between message pairs is the specified feature. For example, we convert the original period to an exclamation mark in the non-standard capitalization example that begins with \textit{just kidding!}, since retaining a period when using standard capitalization in the more formal example, \textit{Just kidding.}, may cause a tonal difference that distracts from the main purpose of the experiment. 

\subsection{Crowdsourcing design}

For crowdsourcing, we chose a subset of two messages for each stylistic feature and one message that combines multiple features (Table~\ref{tbl:style_baseline}). 

\begin{figure}[t]
\centering
\includegraphics[width=0.4\columnwidth]{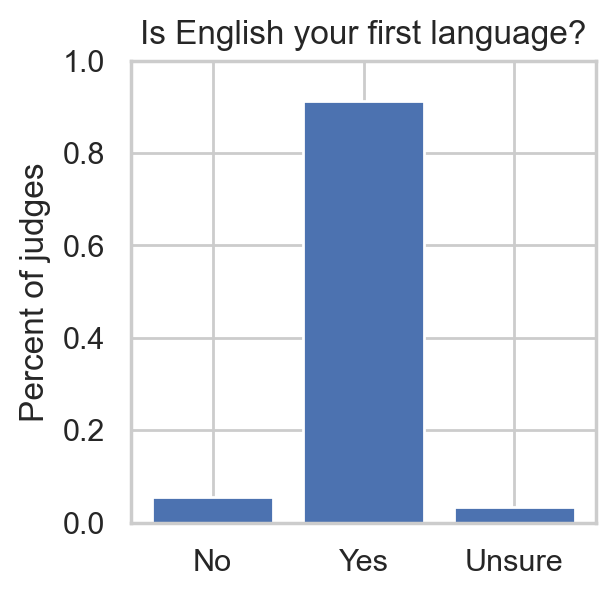}
\includegraphics[width=0.4\columnwidth]{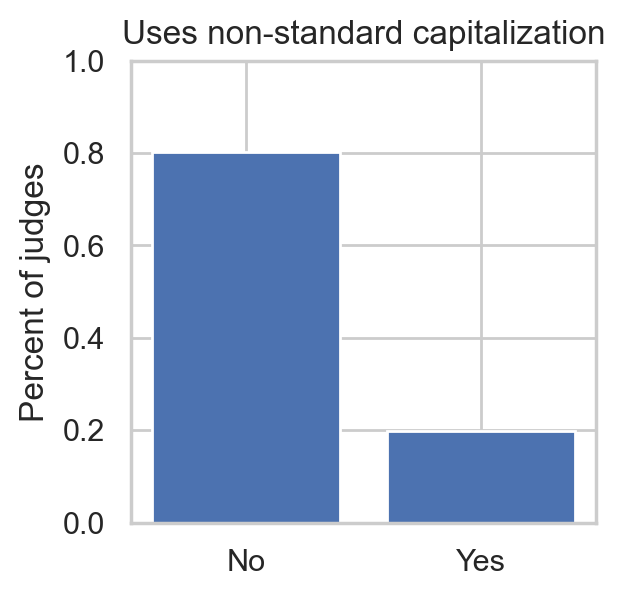}
\includegraphics[width=0.4\columnwidth]{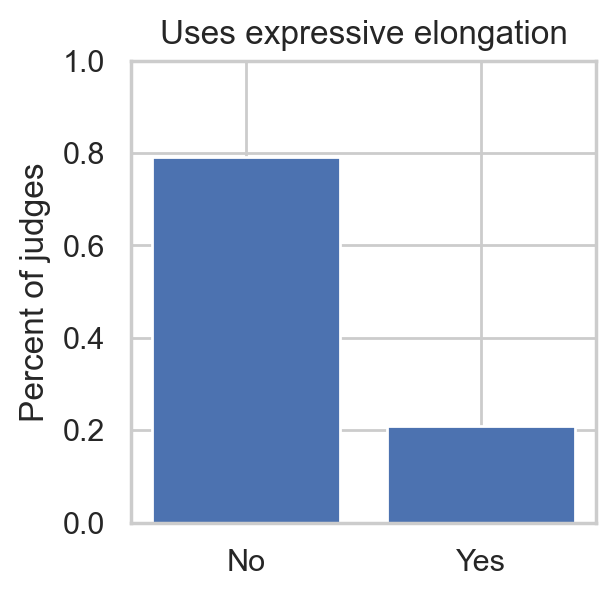}
\includegraphics[width=0.4\columnwidth]{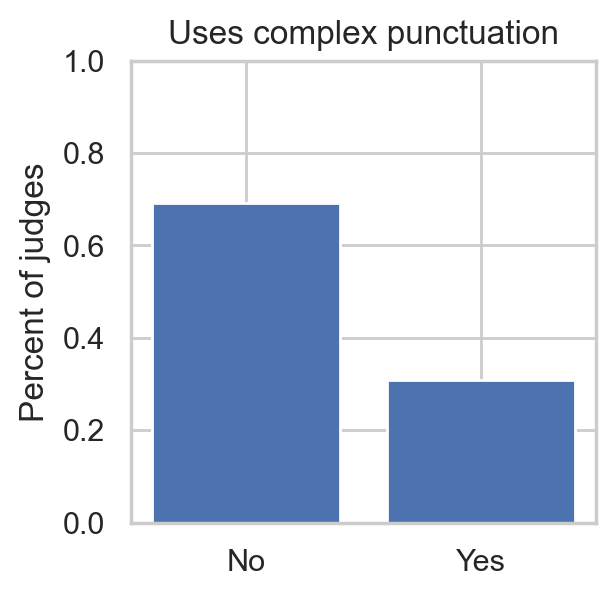}
\caption{Judges' language backgrounds in CS5 ($N=91$).}
\label{fig:CS5_speakers}
\end{figure}

The instructions for this task are same as CS4 (dialects), and the body of this task matches previous case studies. The background questions for this case study are the following: 
\begin{compactitem}
    \item \textit{How many replies did we ask you to compare in this task?} Single-choice options: 1, 2, 3, 4, 5 in randomized order. This is an attention check, where the correct answer is \textit{2}.
    \item \textit{Should reply suggestion systems suggest different replies based on the writing style used in the message?} Single-choice options: Never, Sometimes, Always (Figure~\ref{fig:invariance}). 
    \item \textit{Briefly explain why a reply suggestion system should or should not suggest different responses based on the writing style used in the message.} Free response box.
    \item \textit{When you write emails, do you use any of the following features? Check all that apply.} Options: \textit{lengthening words for emphasis (e.g., writing ``cool'' as ``coooool'')}; \textit{non-standard capitalization (e.g., writing ``I'' as ``i'' or writing words in all lowercase or all caps)}; \textit{complex punctuation (e.g., repeating and/or combining ``?'' and ``!'' like in ``What???!'' or ``Hi!!!'')}, \textit{none of the above} (Figure~\ref{fig:CS5_speakers}).
    \item \textit{Do you speak English as your first language?} Single-choice options: No, I don't; Yes, I do; Unsure (Figure~\ref{fig:CS5_speakers}). 
    \item \textit{(Optional) Please provide us feedback on this task, such as questions that were confusing or unclear.} Free response box.
\end{compactitem}

\begin{figure}[t]
\centering
\includegraphics[trim= 0 2 0 4, clip, width=0.9\columnwidth]{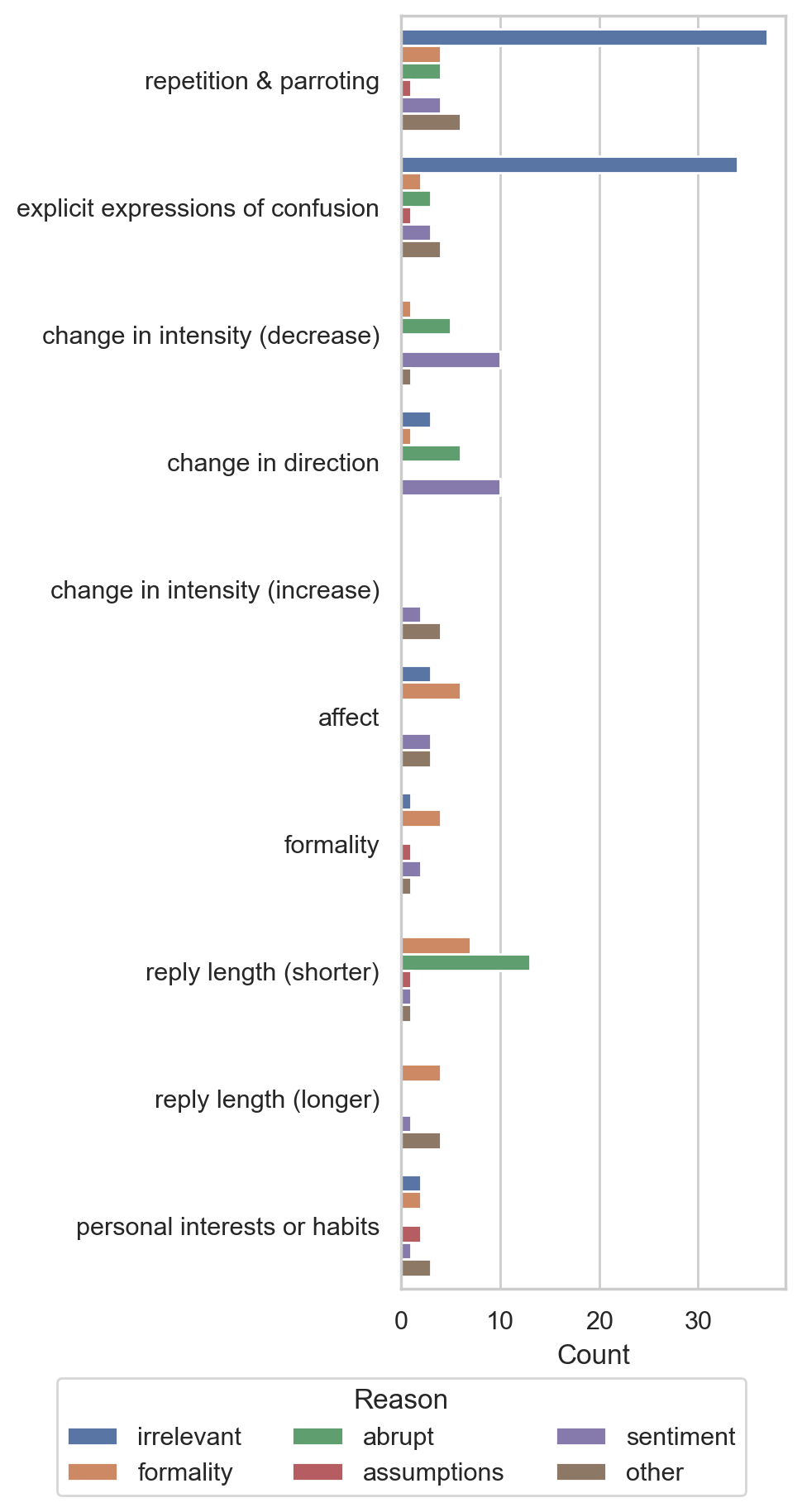}
\caption{Reasons judges marked the second reply as less usable or not usable in CS5. The second reply differs from the baseline reply option along the subcategory of reply behavior shown on the $y$-axis.}
\vspace{-5pt}
\label{fig:CS5_reasons}
\end{figure}

\begin{figure}[t]
\centering
\includegraphics[width=0.45\columnwidth]{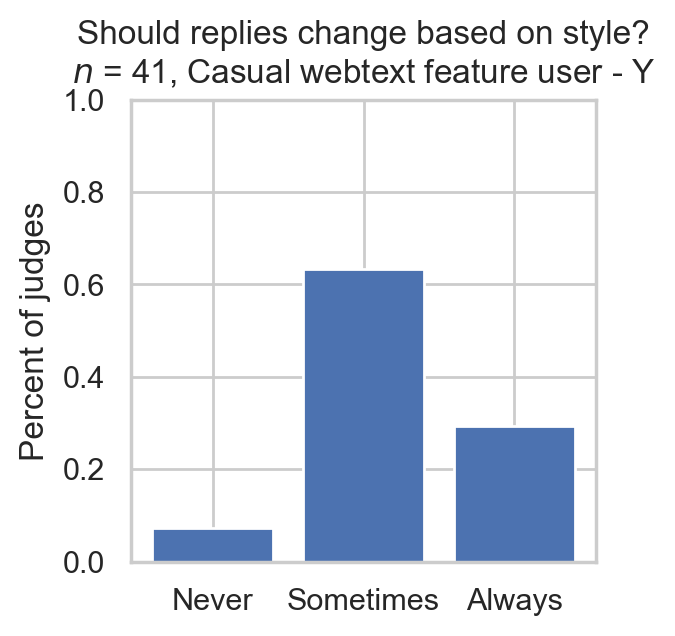}
\includegraphics[width=0.45\columnwidth]{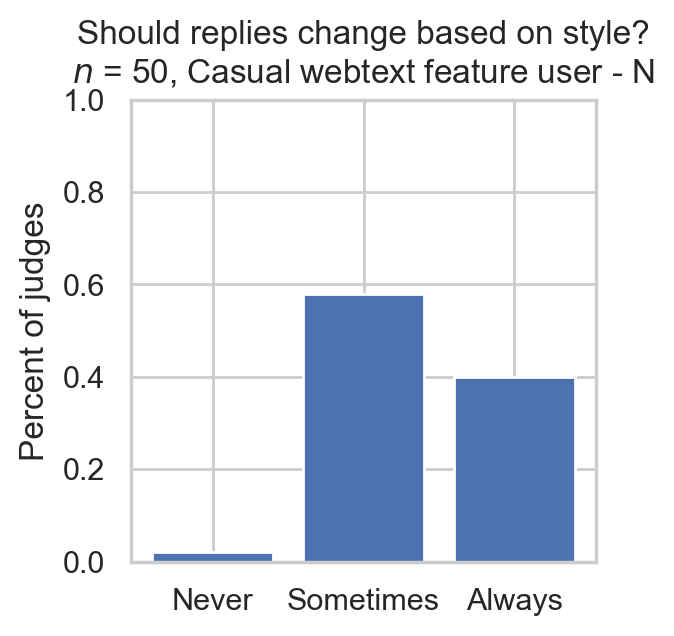}
\caption{Beliefs around whether replies should vary in response to style may shift depending on speakers' own feature use.}
\label{fig:CS5_background}
\end{figure}

\begin{figure*}
\centering
\includegraphics[width=0.9\textwidth]{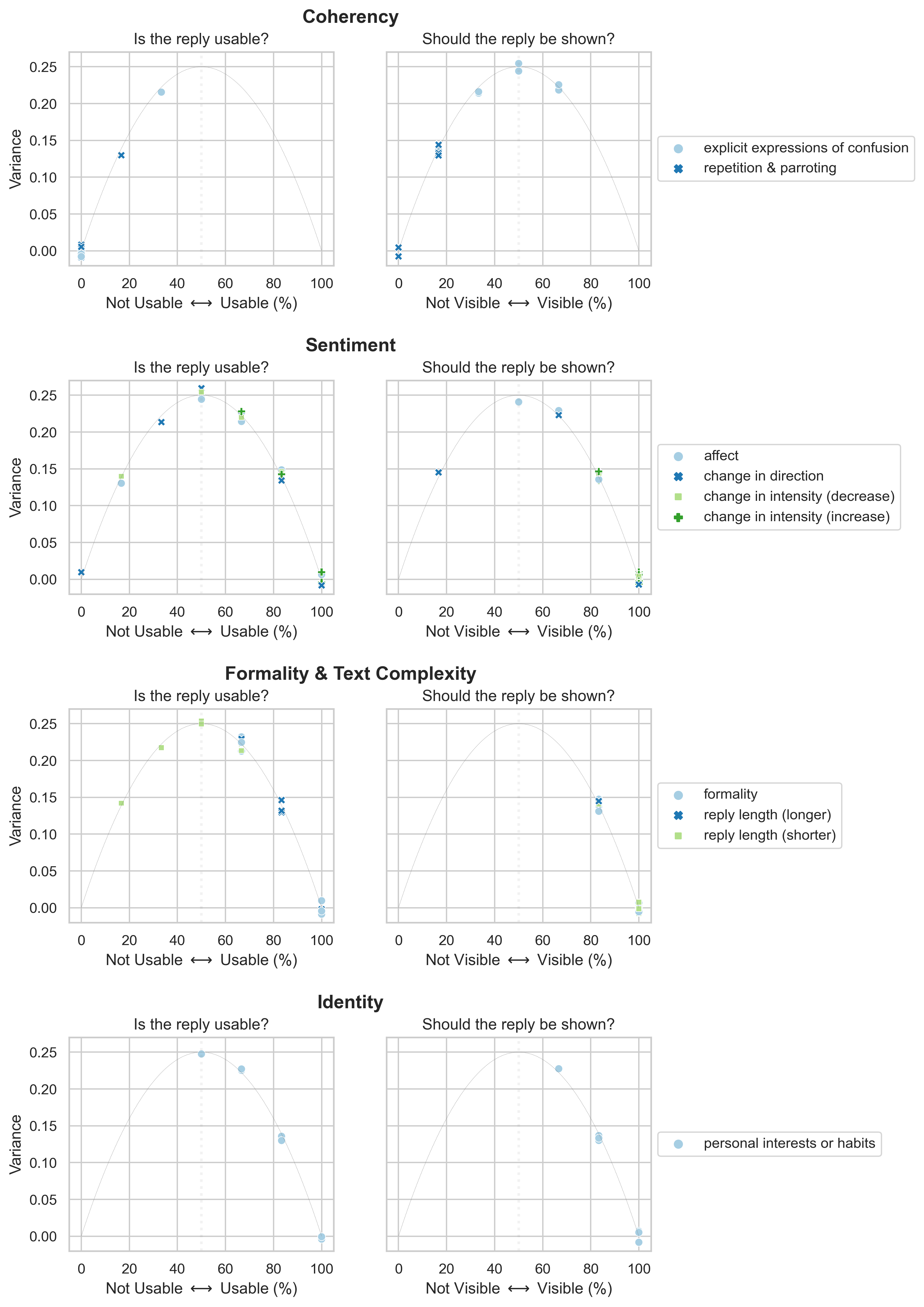}
\caption{In these plots, each point is a message template, and the probability ($x$-axis) that a second reply option is usable (left) and visible (right) is aggregated across the two variants of that message (CS5). When there is less consensus around a reply behavior, variance ($y$-axis) is high. A light vertical gray line emphasizes the highest possible variance, and jitter is added along the $y$-axis so that overlapping points are more visible.}
\label{fig:CS5_variance}
\end{figure*}

\subsection{Crowdsourcing results}

\para{Reply pair validity.} Figure~\ref{fig:CS5_reasons} shows the frequency of various reasons being marked by judges as less usable or unusable modified replies. Typically, the most common reason matched the intentions of our design. Like in CS4, replies involving personal interests or habits in CS5 were not perceived as assumptious as the same subcategory in CS2--3. 

\para{Responses to background questions.} Complex punctuation use is more common than expressive elongation and non-standard capitalization among judges, and 45.05\% of judges use any of the informal-web-text features we tested (Figure~\ref{fig:CS5_speakers}). In addition, judges in CS5 who use these informal-web-text features are slightly less likely to favor systems adapting to messages' language style (Figure~\ref{fig:CS5_background}).

\para{Judges' edited replies.} As described in the main text (\S\ref{expected}), some judges advocated for replies that accommondated, or ``\textit{matched}'', the style of the message. Stylistic accommodation can be tricky to identify, as some judges edit replies across CS1--5 with nonstandard capitalization, especially in all lower case, and without ``proper'' punctuation. Occasionally in CS5, judges crafted replies to messages, especially the message about South Florida that combined multiple features, with a mix of all-uppercase and all-lowercase words, and complex punctuation. 

\para{Aggregated reply preferences.} Figure~\ref{fig:CS5_variance} shows probabilities of reply usability and visibility across message templates. Like in CS1--4, we find that the reply containing an explicit expression of confusion has the highest variance around its visibility, which suggests that clarification requests are not always interpreted as a system's failure to understand a message. Like in CS4, assumptions around personal interests were considered mostly usable in some scenarios, likely because this subcategory was designed similarly across CS4--5. 